\documentclass{article}

\PassOptionsToPackage{numbers, compress}{natbib}

\usepackage[preprint]{neurips_2022}
\newcommand{\finalcopy}{\cvprfinalcopy}
\usepackage[dvipsnames,svgnames,x11names]{xcolor}
\usepackage{tikz}
\usetikzlibrary{arrows.meta,shapes,calc,matrix,fit,positioning,backgrounds,decorations.markings,fadings}
\usepackage{pgfplots}
\usepackage{pgfplotstable}
\pgfplotsset{compat=1.9}
\usepackage{xstring}

\usepgfplotslibrary{external}

\IfBeginWith*{\jobname}{fig/extern/}{\finalcopy}{}

\tikzstyle{every picture}+=[
	remember picture,
	every text node part/.style={align=center},
	every matrix/.append style={ampersand replacement=\&},
]
\tikzstyle{tight} = [inner sep=0pt,outer sep=0pt]
\tikzstyle{node}  = [draw,circle,tight,minimum size=12pt,anchor=center]
\tikzstyle{op}    = [draw,circle,tight]
\tikzstyle{dot}   = [fill,draw,circle,inner sep=1pt,outer sep=0]
\tikzstyle{pt}    = [fill,draw,circle,inner sep=1.5pt,outer sep=.2pt]
\tikzstyle{box}   = [draw,rectangle,inner sep=3pt]
\tikzstyle{high}  = [black!60]
\tikzstyle{group} = [high,box,opacity=.5]
\tikzstyle{dim1}  = [fill opacity=.3,text opacity=1]
\tikzstyle{dim2}  = [fill opacity=.5,text opacity=1]
\tikzstyle{dim3}  = [fill opacity=.7,text opacity=1]
\tikzstyle{rectc} = [tight,transform shape]
\tikzstyle{rect}  = [rectc,anchor=south west]

\newcommand{\leg}[1]{\addlegendentry{#1}}

\tikzset{every mark/.append style={solid}}
\pgfplotsset{
	grid=both, width=\columnwidth, try min ticks=5,
	every axis/.append style={font=\small},
	every axis plot/.append style={thick,mark=none,mark size=1.8,tension=0.18},
	legend cell align=left, legend style={fill opacity=0.8},
	xticklabel={\pgfmathprintnumber[assume math mode=true]{\tick}},
	yticklabel={\pgfmathprintnumber[assume math mode=true]{\tick}},
	nodes near coords math/.style={
		nodes near coords={\pgfmathprintnumber[assume math mode=true]{\pgfplotspointmeta}},
	},
}

\pgfplotsset{
	dash/.style={mark=o,dashed,opacity=0.6},
	dott/.style={mark=o,dotted,opacity=0.6},
	nolim/.style={enlargelimits=false},
	plain/.style={every axis plot/.append style={},nolim,grid=none},
}

\pgfdeclarelayer{bg4}
\pgfdeclarelayer{bg3}
\pgfdeclarelayer{bg2}
\pgfdeclarelayer{bg1}
\pgfdeclarelayer{fg1}
\pgfdeclarelayer{fg2}
\pgfdeclarelayer{fg3}
\pgfdeclarelayer{fg4}
\pgfsetlayers{bg4,bg3,bg2,bg1,main,fg1,fg2,fg3,fg4}

\pgfkeys{/tikz/.cd, aspect/.store in=\aspect, aspect=1}
\pgfkeys{/tikz/.cd, depth/.store in=\depth, depth=.5}
\pgfkeys{/tikz/.cd, stepx/.store in=\step, stepx=1}

\tikzstyle{geom} = [line join=bevel,aspect=1,depth=.5,z={(\depth*\aspect,\depth)}]
\tikzstyle{wire} = [geom,draw,thick]

\def\cx[#1,#2,#3]{#1}
\def\cy[#1,#2,#3]{#2}
\def\cz[#1,#2,#3]{#3}
\def\ex[#1,#2,#3]{#1,0,0}
\def\ey[#1,#2,#3]{0,#2,0}
\def\ez[#1,#2,#3]{0,0,#3}

\newcommand{\zrect}[3][]{%
\path[geom,#1] #2 rectangle +(\cx[#3],\cy[#3]);
}
\newcommand{\yrect}[3][]{%
\path[geom,#1,shift={#2},xslant=\aspect]
	(0,0) rectangle +(\cx[#3],\depth*\cz[#3]);
}
\newcommand{\xrect}[3][]{%
\path[geom,#1,shift={#2},yslant=1/\aspect]
	(0,0) rectangle +(\aspect*\depth*\cz[#3],\cy[#3]);
}

\makeatletter
\renewcommand\paragraph{\@startsection{paragraph}{4}{\z@}{1ex}{-1em}{\normalfont\normalsize\bfseries}}
\makeatother

\usepackage{times}
\usepackage{epsfig}
\usepackage{authblk}
\usepackage{graphicx}
\usepackage{amsmath}
\usepackage{amssymb}
\usepackage[utf8]{inputenc}
\usepackage{mathtools}
\usepackage[dvipsnames]{xcolor}
\usepackage{color, colortbl}
\usepackage{enumitem}
\usepackage{xspace}
\usepackage{bm}
\usepackage{array,booktabs}
\usepackage{makecell}
\usepackage{multirow}
\usepackage[normalem]{ulem}
\usepackage{verbatim}
\usepackage{kantlipsum}
\usepackage{wrapfig}
\usepackage{comment}
\usepackage{float}

\usepackage[pagebackref,breaklinks,colorlinks]{hyperref}
\usepackage[ruled,vlined,linesnumbered]{algorithm2e}

\usepackage{caption}
\usepackage{subcaption}

\newcommand{\ewa}[1]{}
\newcommand{\sv}[1]{}
\newcommand{\iavr}[1]{}
\newcommand{\la}[1]{}

\newcommand{\printfnsymbol}[1]{%
  \textsuperscript{\@fnsymbol{#1}}%
}

\title{Teach me how to Interpolate a Myriad of Embeddings}

\author{Shashanka Venkataramanan$^1$  \hspace{1.5em}  Ewa Kijak$^1$ \hspace{1.5em}  Laurent Amsaleg$^1$ \hspace{1.5em} Yannis Avrithis$^2$\\
\vspace{6pt}
$^1$Inria, Univ Rennes, CNRS, IRISA \\ $^2$Institute of Advanced Research in Artificial Intelligence (IARAI), Athena RC\\
}

\begin{document}

\maketitle


\newcommand{\head}[1]{{\smallskip\noindent\textbf{#1}}}
\newcommand{\alert}[1]{{\color{red}{#1}}}
\newcommand{\sm}{\scriptsize}
\newcommand{\eq}[1]{(\ref{eq:#1})}

\newcommand{\Th}[1]{\textsc{#1}}
\newcommand{\mr}[2]{\multirow{#1}{*}{#2}}
\newcommand{\mc}[2]{\multicolumn{#1}{c}{#2}}
\newcommand{\tb}[1]{\textbf{#1}}
\newcommand{\ul}[1]{\underline{#1}}
\newcommand{\ch}{\checkmark}

\newcommand{\red}[1]{{\color{red}{#1}}}
\newcommand{\blue}[1]{{\color{blue}{#1}}}
\newcommand{\green}[1]{\color{green}{#1}}
\newcommand{\gray}[1]{{\color{gray}{#1}}}

\newcommand{\citeme}[1]{\red{[XX]}}
\newcommand{\refme}[1]{\red{(XX)}}

\newcommand{\fig}[2][1]{\includegraphics[width=#1\linewidth]{fig/#2}}
\newcommand{\figh}[2][1]{\includegraphics[height=#1\linewidth]{fig/#2}}


\newcommand{\tran}{^\top}
\newcommand{\mtran}{^{-\top}}
\newcommand{\zcol}{\mathbf{0}}
\newcommand{\zrow}{\zcol\tran}

\newcommand{\ind}{\mathbbm{1}}
\newcommand{\expect}{\mathbb{E}}
\newcommand{\nat}{\mathbb{N}}
\newcommand{\zahl}{\mathbb{Z}}
\newcommand{\real}{\mathbb{R}}
\newcommand{\proj}{\mathbb{P}}
\newcommand{\prob}{\mathbf{Pr}}
\newcommand{\normal}{\mathcal{N}}

\newcommand{\mif}{\textrm{if}\ }
\newcommand{\other}{\textrm{otherwise}}
\newcommand{\minimize}{\textrm{minimize}\ }
\newcommand{\maximize}{\textrm{maximize}\ }
\newcommand{\st}{\textrm{subject\ to}\ }

\newcommand{\id}{\operatorname{id}}
\newcommand{\const}{\operatorname{const}}
\newcommand{\sgn}{\operatorname{sgn}}
\newcommand{\var}{\operatorname{Var}}
\newcommand{\mean}{\operatorname{mean}}
\newcommand{\trace}{\operatorname{tr}}
\newcommand{\diag}{\operatorname{diag}}
\newcommand{\vect}{\operatorname{vec}}
\newcommand{\cov}{\operatorname{cov}}
\newcommand{\sign}{\operatorname{sign}}
\newcommand{\prj}{\operatorname{proj}}

\newcommand{\softmax}{\operatorname{softmax}}
\newcommand{\clip}{\operatorname{clip}}

\newcommand{\defn}{\mathrel{:=}}
\newcommand{\peq}{\mathrel{+\!=}}
\newcommand{\meq}{\mathrel{-\!=}}

\newcommand{\floor}[1]{\left\lfloor{#1}\right\rfloor}
\newcommand{\ceil}[1]{\left\lceil{#1}\right\rceil}
\newcommand{\inner}[1]{\left\langle{#1}\right\rangle}
\newcommand{\norm}[1]{\left\|{#1}\right\|}
\newcommand{\abs}[1]{\left|{#1}\right|}
\newcommand{\frob}[1]{\norm{#1}_F}
\newcommand{\card}[1]{\left|{#1}\right|\xspace}
\newcommand{\divg}[2]{{#1\ ||\ #2}}
\newcommand{\diff}{\mathrm{d}}
\newcommand{\der}[3][]{\frac{d^{#1}#2}{d#3^{#1}}}
\newcommand{\pder}[3][]{\frac{\partial^{#1}{#2}}{\partial{#3^{#1}}}}
\newcommand{\ipder}[3][]{\partial^{#1}{#2}/\partial{#3^{#1}}}
\newcommand{\dder}[3]{\frac{\partial^2{#1}}{\partial{#2}\partial{#3}}}

\newcommand{\wb}[1]{\overline{#1}}
\newcommand{\wt}[1]{\widetilde{#1}}

\def\xssp{\hspace{1pt}}
\def\ssp{\hspace{3pt}}
\def\msp{\hspace{5pt}}
\def\lsp{\hspace{12pt}}

\newcommand{\cA}{\mathcal{A}}
\newcommand{\cB}{\mathcal{B}}
\newcommand{\cC}{\mathcal{C}}
\newcommand{\cD}{\mathcal{D}}
\newcommand{\cE}{\mathcal{E}}
\newcommand{\cF}{\mathcal{F}}
\newcommand{\cG}{\mathcal{G}}
\newcommand{\cH}{\mathcal{H}}
\newcommand{\cI}{\mathcal{I}}
\newcommand{\cJ}{\mathcal{J}}
\newcommand{\cK}{\mathcal{K}}
\newcommand{\cL}{\mathcal{L}}
\newcommand{\cM}{\mathcal{M}}
\newcommand{\cN}{\mathcal{N}}
\newcommand{\cO}{\mathcal{O}}
\newcommand{\cP}{\mathcal{P}}
\newcommand{\cQ}{\mathcal{Q}}
\newcommand{\cR}{\mathcal{R}}
\newcommand{\cS}{\mathcal{S}}
\newcommand{\cT}{\mathcal{T}}
\newcommand{\cU}{\mathcal{U}}
\newcommand{\cV}{\mathcal{V}}
\newcommand{\cW}{\mathcal{W}}
\newcommand{\cX}{\mathcal{X}}
\newcommand{\cY}{\mathcal{Y}}
\newcommand{\cZ}{\mathcal{Z}}

\newcommand{\vA}{\mathbf{A}}
\newcommand{\vB}{\mathbf{B}}
\newcommand{\vC}{\mathbf{C}}
\newcommand{\vD}{\mathbf{D}}
\newcommand{\vE}{\mathbf{E}}
\newcommand{\vF}{\mathbf{F}}
\newcommand{\vG}{\mathbf{G}}
\newcommand{\vH}{\mathbf{H}}
\newcommand{\vI}{\mathbf{I}}
\newcommand{\vJ}{\mathbf{J}}
\newcommand{\vK}{\mathbf{K}}
\newcommand{\vL}{\mathbf{L}}
\newcommand{\vM}{\mathbf{M}}
\newcommand{\vN}{\mathbf{N}}
\newcommand{\vO}{\mathbf{O}}
\newcommand{\vP}{\mathbf{P}}
\newcommand{\vQ}{\mathbf{Q}}
\newcommand{\vR}{\mathbf{R}}
\newcommand{\vS}{\mathbf{S}}
\newcommand{\vT}{\mathbf{T}}
\newcommand{\vU}{\mathbf{U}}
\newcommand{\vV}{\mathbf{V}}
\newcommand{\vW}{\mathbf{W}}
\newcommand{\vX}{\mathbf{X}}
\newcommand{\vY}{\mathbf{Y}}
\newcommand{\vZ}{\mathbf{Z}}

\newcommand{\va}{\mathbf{a}}
\newcommand{\vb}{\mathbf{b}}
\newcommand{\vc}{\mathbf{c}}
\newcommand{\vd}{\mathbf{d}}
\newcommand{\ve}{\mathbf{e}}
\newcommand{\vf}{\mathbf{f}}
\newcommand{\vg}{\mathbf{g}}
\newcommand{\vh}{\mathbf{h}}
\newcommand{\vi}{\mathbf{i}}
\newcommand{\vj}{\mathbf{j}}
\newcommand{\vk}{\mathbf{k}}
\newcommand{\vl}{\mathbf{l}}
\newcommand{\vm}{\mathbf{m}}
\newcommand{\vn}{\mathbf{n}}
\newcommand{\vo}{\mathbf{o}}
\newcommand{\vp}{\mathbf{p}}
\newcommand{\vq}{\mathbf{q}}
\newcommand{\vr}{\mathbf{r}}
\newcommand{\Vs}{\mathbf{s}}
\newcommand{\vt}{\mathbf{t}}
\newcommand{\vu}{\mathbf{u}}
\newcommand{\vv}{\mathbf{v}}
\newcommand{\vw}{\mathbf{w}}
\newcommand{\vx}{\mathbf{x}}
\newcommand{\vy}{\mathbf{y}}
\newcommand{\vz}{\mathbf{z}}

\newcommand{\vone}{\mathbf{1}}
\newcommand{\vzero}{\mathbf{0}}

\newcommand{\valpha}{{\boldsymbol{\alpha}}}
\newcommand{\vbeta}{{\boldsymbol{\beta}}}
\newcommand{\vgamma}{{\boldsymbol{\gamma}}}
\newcommand{\vdelta}{{\boldsymbol{\delta}}}
\newcommand{\vepsilon}{{\boldsymbol{\epsilon}}}
\newcommand{\vzeta}{{\boldsymbol{\zeta}}}
\newcommand{\veta}{{\boldsymbol{\eta}}}
\newcommand{\vtheta}{{\boldsymbol{\theta}}}
\newcommand{\viota}{{\boldsymbol{\iota}}}
\newcommand{\vkappa}{{\boldsymbol{\kappa}}}
\newcommand{\vlambda}{{\boldsymbol{\lambda}}}
\newcommand{\vmu}{{\boldsymbol{\mu}}}
\newcommand{\vnu}{{\boldsymbol{\nu}}}
\newcommand{\vxi}{{\boldsymbol{\xi}}}
\newcommand{\vomikron}{{\boldsymbol{\omikron}}}
\newcommand{\vpi}{{\boldsymbol{\pi}}}
\newcommand{\vrho}{{\boldsymbol{\rho}}}
\newcommand{\vsigma}{{\boldsymbol{\sigma}}}
\newcommand{\vtau}{{\boldsymbol{\tau}}}
\newcommand{\vupsilon}{{\boldsymbol{\upsilon}}}
\newcommand{\vphi}{{\boldsymbol{\phi}}}
\newcommand{\vchi}{{\boldsymbol{\chi}}}
\newcommand{\vpsi}{{\boldsymbol{\psi}}}
\newcommand{\vomega}{{\boldsymbol{\omega}}}

\newcommand{\rLambda}{\mathrm{\Lambda}}
\newcommand{\rSigma}{\mathrm{\Sigma}}

\newcommand{\vLambda}{\bm{\rLambda}}
\newcommand{\vSigma}{\bm{\rSigma}}


\makeatletter
\newcommand{\vast}[1]{\bBigg@{#1}}
\makeatother

\makeatletter
\newcommand*\bdot{\mathpalette\bdot@{.7}}
\newcommand*\bdot@[2]{\mathbin{\vcenter{\hbox{\scalebox{#2}{$\m@th#1\bullet$}}}}}
\makeatother

\makeatletter
\DeclareRobustCommand\onedot{\futurelet\@let@token\@onedot}
\def\@onedot{\ifx\@let@token.\else.\null\fi\xspace}

\def\eg{\emph{e.g}\onedot} \def\Eg{\emph{E.g}\onedot}
\def\ie{\emph{i.e}\onedot} \def\Ie{\emph{I.e}\onedot}
\def\cf{\emph{cf}\onedot} \def\Cf{\emph{Cf}\onedot}
\def\etc{\emph{etc}\onedot} \def\vs{\emph{vs}\onedot}
\def\wrt{w.r.t\onedot} \def\dof{d.o.f\onedot} \def\aka{a.k.a\onedot}
\def\etal{\emph{et al}\onedot}
\makeatother

\newcommand{\relu}{\operatorname{relu}}
\newcommand{\mix}{\operatorname{mix}}
\newcommand{\Mix}{\operatorname{Mix}}
\newcommand{\Beta}{\operatorname{Beta}}
\newcommand{\Dir}{\operatorname{Dir}}

\newcommand{\gp}[1]{{\color{ForestGreen}#1}}
\newcommand{\sota}[1]{\red{\textbf{#1}}}
\newcommand{\negative}[1]{\blue{#1}}

\definecolor{LightCyan}{rgb}{0.88,1,1}

\newcommand{\gn}[1]{\red{#1}}
\newcommand{\se}[1]{\blue{#1}}

\begin{abstract}
\emph{Mixup} refers to interpolation-based data augmentation, originally motivated as a way to go beyond \emph{empirical risk minimization} (ERM). Yet, its extensions focus on the definition of interpolation and the space where it takes place, while the augmentation itself is less studied: For a mini-batch of size $m$, most methods interpolate between $m$ pairs with a single scalar interpolation factor $\lambda$.

In this work, we make progress in this direction by introducing \emph{MultiMix}, which interpolates an arbitrary number $n$ of tuples, each of length $m$, with one vector $\lambda$ per tuple. On sequence data, we further extend to \emph{dense} interpolation and loss computation over all spatial positions. Overall, we increase the number of tuples per mini-batch by orders of magnitude at little additional cost. This is possible by interpolating at the very last layer before the classifier. Finally, to address inconsistencies due to linear target interpolation, we introduce a \emph{self-distillation} approach to generate and interpolate synthetic targets.

We empirically show that our contributions result in significant improvement over state-of-the-art mixup methods on four benchmarks. By analyzing the embedding space, we observe that the classes are more tightly clustered and uniformly spread over the embedding space, thereby explaining the improved behavior.
\end{abstract}

\section{Introduction}
\label{sec:intro}

\emph{Mixup}~\cite{zhang2018mixup} is a data augmentation method that interpolates between pairs of training examples, thus regularizing a neural network to favor linear behavior in-between examples. Besides improving generalization, it has important properties such as reducing overconfident predictions and increasing the robustness to adversarial examples. Several follow-up works have studied interpolation in the \emph{latent} or \emph{embedding} space, which is equivalent to interpolating along a manifold in the input space~\cite{verma2019manifold}, and a number of nonlinear and attention-based interpolation mechanisms~\cite{yun2019cutmix, kim2020puzzle, kim2021co, uddin2020saliencymix, hong2021stylemix}. However, little progress has been made in the augmentation process itself, \ie, the number of examples being interpolated and the number of interpolated examples being generated.

\begin{figure}
\small
\centering
\setlength{\tabcolsep}{20pt}
\newcommand{\sz}{.3}
\begin{tabular}{cc}
\fig[\sz]{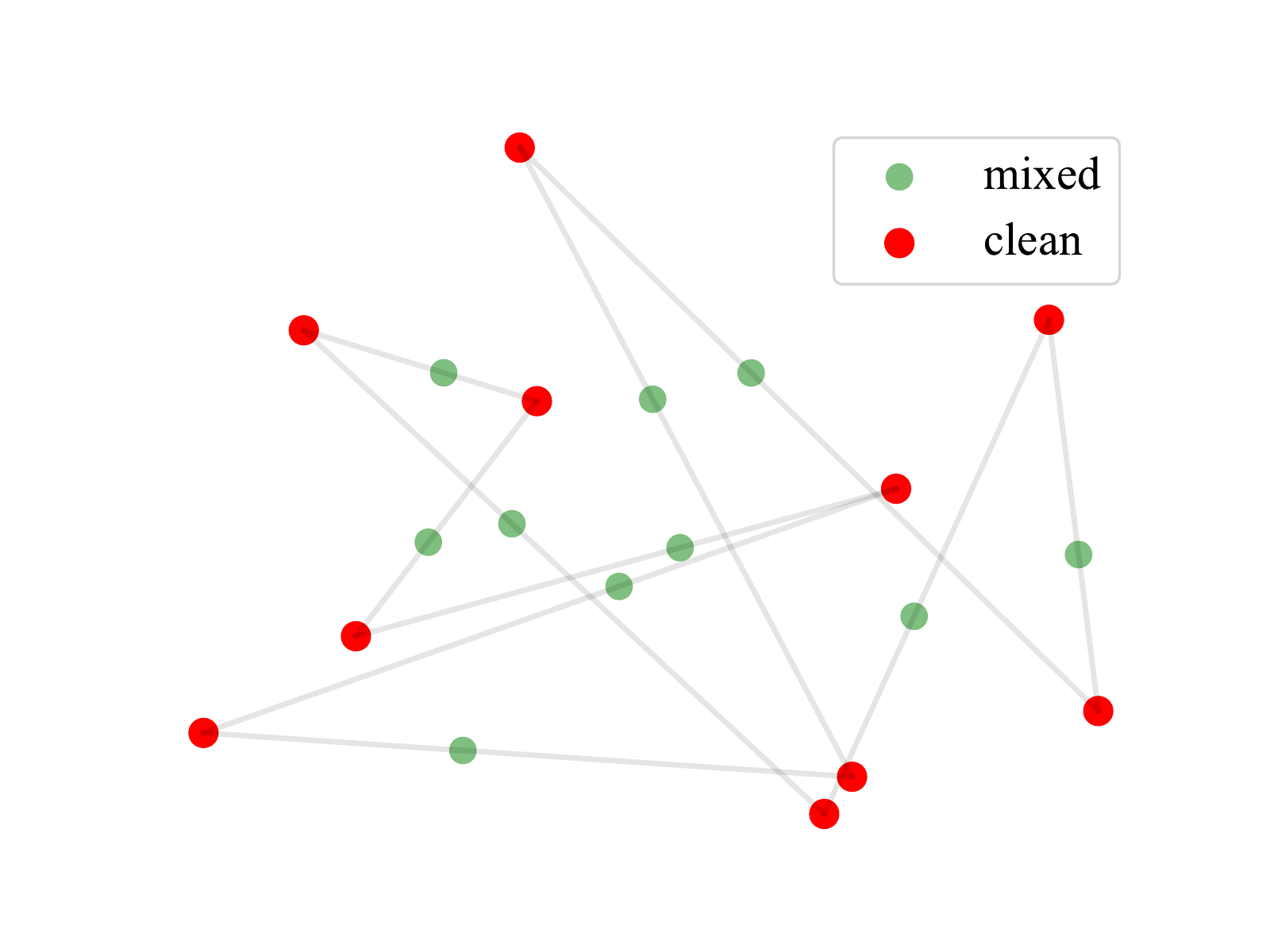}  &
\fig[\sz]{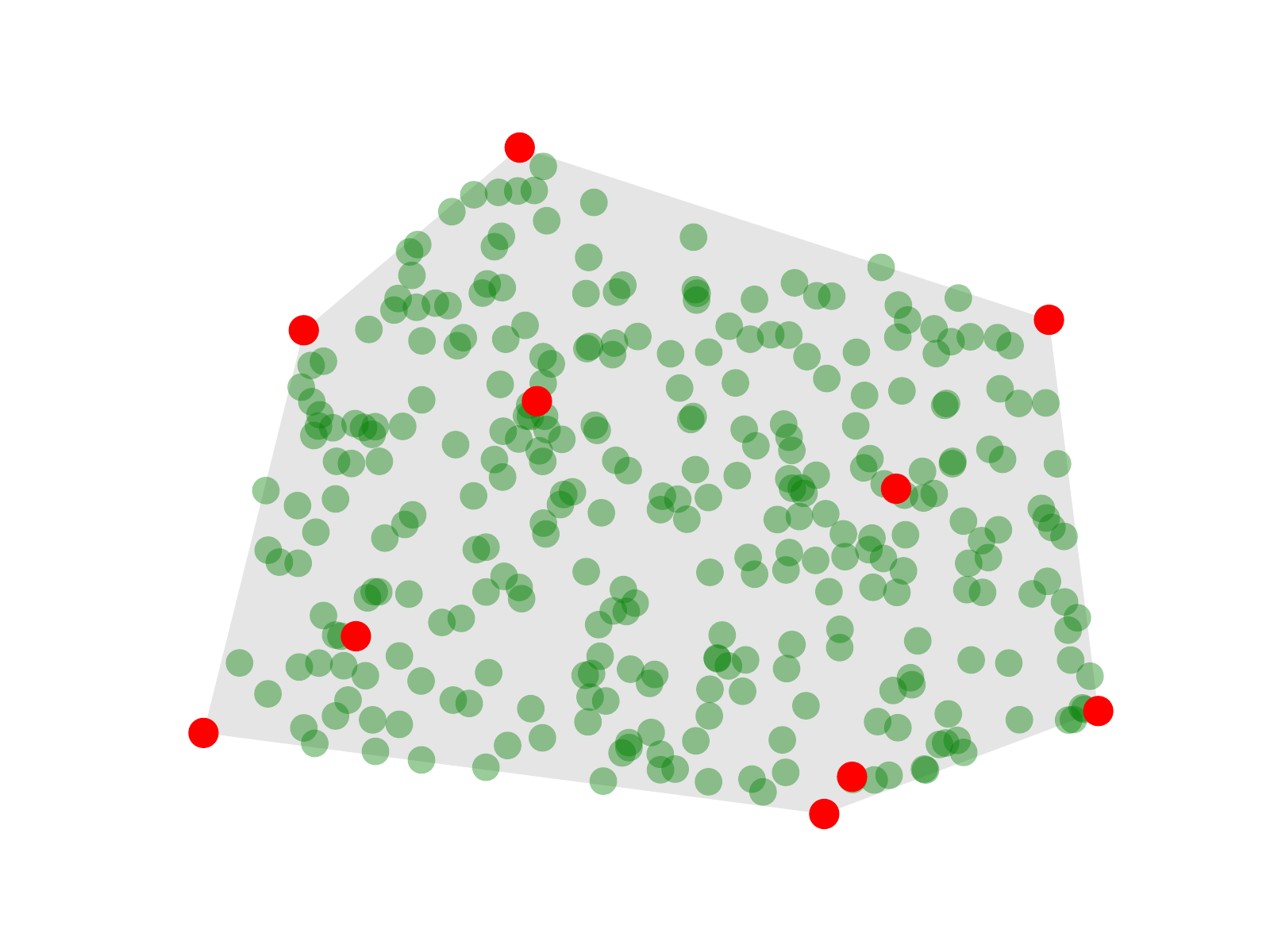} \\
(a) mixup &
(b) MultiMix (ours)
\end{tabular}

\caption{Data augmentation from a mini-batch $B$ consisting of $m=10$ points in two dimensions. (a) \emph{mixup}: sampling of $m$ points on linear segments between $m$ pairs of points in $B$, using the same interpolation factor $\lambda$. (b) \emph{MultiMix}: sampling of $n=300$ points in the convex hull of $B$.}
\label{fig:idea}
\end{figure}

Mixup was originally motivated as a way to go beyond \emph{empirical risk minimization} (ERM)~\cite{Vapn99} through a vicinal distribution expressed as an expectation over an interpolation factor $\lambda$, which is equivalent to the set of linear segments between all pairs of training inputs and targets. In practice however, in every training iteration, a single scalar $\lambda$ is drawn and the number of interpolated pairs is limited to the size of the mini-batch, as illustrated in~\autoref{fig:idea}(a). This is because, if interpolation takes place in the input space, it would be expensive to increase the number of examples per iteration. To our knowledge, these limitations exist in all mixup methods.

In this work, we argue that a data augmentation process should augment the data seen by the model, or at least by its last few layers, as much as possible. In this sense, we follow \emph{manifold mixup}~\cite{verma2019manifold} and generalize it in a number of ways to introduce \emph{MultiMix}, as illustrated in~\autoref{fig:idea}(b). First,
rather than pairs, we interpolate \emph{tuples} that are as large as the mini-batch.
Effectively, instead of linear segments between pairs of examples in the mini-batch, we sample on their entire \emph{convex hull}. Second, we draw a different vector $\lambda$ for each tuple. Third, and most important, we increase the number of interpolated tuples per iteration by orders of magnitude by only slightly decreasing the actual training throughput in examples per second. This is possible by interpolating at the deepest layer possible, \ie, just before the classifier, which also happens to be the most effective choice. The interpolated embeddings are thus only processed by a single layer.

Apart from increasing the number of examples seen by the model, another idea is to increase the number of loss terms per example. In many modalities of interest, the input is a \emph{sequence} in one or more dimensions:
pixels or patches in images, voxels in video, points or triangles in high-dimensional surfaces, to name a few. The structure of input data is expressed in matrices or tensors, which often preserve a certain spatial resolution until the deepest network layer before they collapse \eg by global average pooling~\cite{SLJ+15, he2016deep} or by taking the output of a classification token~\cite{VSP+17, dosovitskiy2020image}.

In this sense, we choose to operate at the level of \emph{sequence elements} rather than representing examples by a single vector. We introduce \emph{dense MultiMix}, which is the first approach of this kind in mixup-based data augmentation. In particular, we interpolate densely the embeddings and targets of sequence elements and we also apply the loss densely, as illustrated in~\autoref{fig:dense}. This is an extreme form of augmentation where the number of interpolated tuples and loss terms increases further by one or two orders of magnitude, but at little cost.

Finally, linear interpolation of targets, which is the norm in most mixup variants, has a limitation: Given two examples with different class labels, the interpolated example may actually lie in a region associated with a third class in the feature space, which is identified as \emph{manifold intrusion}~\cite{guo2019mixup}. In the absence of any data other than the mini-batch, a straightforward way to address this limitation is to devise targets originating in the network itself. This naturally leads to \emph{self-distillation}, whereby a moving average of the network acts as a teacher and provides synthetic soft targets~\cite{tarvainen2017mean}, to be interpolated exactly like the original hard targets.

In summary, we make the following contributions:
\begin{enumerate}[itemsep=2pt, parsep=0pt, topsep=0pt]
	\item We introduce \emph{MultiMix}, which, given a mini-batch of size $m$, interpolates an arbitrary number $n \gg m$ of tuples, each of length $m$, with one interpolation vector $\lambda$ per tuple---compared with $m$ pairs, all with the same scalar $\lambda$ for most mixup methods (\autoref{sec:multi}).
	\item We extend to \emph{dense} interpolation and loss computation over all spatial positions (\autoref{sec:dense}).
	\item We use \emph{online self-distillation} to generate and interpolate soft targets for mixup---compared with linear target interpolation for most mixup methods (\autoref{sec:distill}).
	\item We improve over state-of-the-art mixup methods on \emph{image classification}, \emph{robustness to adversarial attacks}, \emph{object detection} and \emph{out-of-distribution detection}.
	(\autoref{sec:exp}).
\end{enumerate}

\section{Related Work}
\label{sec:related}

\paragraph{Mixup}

In general, mixup interpolates between pairs of input examples~\cite{zhang2018adversarial} or embeddings~\cite{verma2019manifold} and their corresponding target labels. Several follow-up methods mix input images according to spatial position, either at random rectangles~\cite{yun2019cutmix} or based on attention~\cite{uddin2020saliencymix, kim2020puzzle, kim2021co}, in an attempt to focus on a different object in each image. We also use attention in our dense MultiMix variant, but in the embedding space. Other definitions of interpolation include the combination of content and style from two images~\cite{hong2021stylemix} and the spatial alignment of dense features~\cite{venkataramanan2021alignmix}. Our dense MultiMix variant also uses dense features but without aligning them, hence it can mix a very large number of images and generate even more interpolated data.

Our work is orthogonal to these methods as we focus on the sampling process of augmentation rather than on the definition of interpolation. As far as we are aware, the only methods that mix more than two examples are OptTransMix~\cite{zhu2020automix}, which involves a complex optimization process in the input space and only applies to images with clean background, and SuperMix~\cite{Dabouei_2021_CVPR}, which uses a Dirichlet distribution like we do, but interpolates in the input space not more than 3 images, while we interpolate all embeddings of a mini-batch.


\paragraph{Self-distillation}

Distillation refers to a two-stage knowledge transfer process where a larger teacher model or ensemble is trained before predicting soft targets to train a smaller student model on the same~\cite{bucilua2006model, hinton2015distilling, romero2014fitnets, shen2019meal, zagoruyko2016paying} or different~\cite{RDG+18, xie2020self} training data. The architecture of the two models may be the same with training at multiple stages, for example in continual learning~\cite{rebuffi2017icarl, LiHo18}. In self-distillation or co-distillation, not only the models are the same, but the knowledge transfer process is also online, \eg between layers of the same model~\cite{zhang2019your} or between two versions of the model~\cite{anil2018large}, where the teacher parameters may be obtained from the student rather than learned~\cite{tarvainen2017mean}. The latter approach has been successful in self-supervised representation learning~\cite{grill2020bootstrap, caron2021emerging, zhou2021ibot}. As far as we know, distillation has only been used for mixup as a two stage process between different models~\cite{Dabouei_2021_CVPR} and we are the first to use online self-distillation in this context, following~\cite{tarvainen2017mean}.


\paragraph{Dense loss functions}

Although standard in dense tasks like semantic segmentation~\cite{noh2015learning, he2017mask}, where dense targets commonly exist, dense loss functions are less common otherwise. Few examples are in few-shot learning~\cite{Lifchitz_2019_CVPR, Li_2019_CVPR}, where data augmentation is of utter importance, and in unsupervised representation learning, \eg dense contrastive learning~\cite{o2020unsupervised, Wang_2021_CVPR}, learning from spatial correspondences~\cite{xiong2021self, xie2021propagate} and masked language or image modeling~\cite{bert, xie2021simmim, li2021mst, zhou2021ibot}. Some of these methods use dense distillation~\cite{xiong2021self, zhou2021ibot}, which is also studied in continual  learning~\cite{Dhar_2019_CVPR, DCO+20}. To our knowledge, we are the first to use dense interpolation and a dense loss function for mixup. Our setting is supervised, similar to dense classification~\cite{Lifchitz_2019_CVPR}, but we also use dense distillation~\cite{zhou2021ibot}.

\section{Method}
\label{sec:method}

\subsection{Preliminaries and background}

\paragraph{Problem formulation}

Let $x \in \cX$ be an input example and $y \in \cY$ its one-hot encoded target, where $\cX = \real^D$ is the input space, $\cY = \{0, 1\}^c$ and $c$ is the total number of classes. Let $f_\theta: \cX \to \real^{d}$ be an encoder that maps the input $x$ to an embedding $z = f_\theta(x)$, where $d$ is the dimension of the embedding. A classifier $g_W: \real^{d} \to \Delta^{c-1}$  maps $z$ to a vector $p = g_W(z)$ of predicted probabilities over classes, where $\Delta^n \subset \real^{n+1}$ is the unit $n$-simplex, \ie, $p \ge 0$ and $\vone_c\tran p = 1$, and $\vone_c \in \real^c$ is an all-ones vector. The overall network mapping is $f \defn g_W \circ f_\theta$.

Parameters $(\theta, W)$ are learned by optimizing over mini-batches. Given a mini-batch of $m$ examples, let $X = (x_1, \dots, x_m) \in \real^{D \times m}$ be the inputs, $Y = (y_1, \dots, y_m) \in \real^{c \times m}$ the targets and $P = (p_1, \dots, p_m) \in \real^{c \times m}$ the predicted probabilities of the mini-batch, where $P = f(X) \defn (f(x_1), \dots, f(x_m))$. The objective is to minimize the cross-entropy
\begin{align}
	H(Y, P) \defn - \vone_c\tran (Y \odot \log(P)) \vone_m / m
\label{eq:ce}
\end{align}
of predicted probabilities $P$ relative to targets $Y$ averaged over the mini-batch, where $\odot$ is the Hadamard (element-wise) product. In summary, the mini-batch loss is
\begin{align}
	L(X,Y;\theta,W) \defn H(Y, g_W(f_\theta(X))).
\label{eq:loss}
\end{align}

\paragraph{Mixup}

Mixup methods commonly interpolate pairs of inputs or embeddings and the corresponding targets at the mini-batch level while training. Given a mini-batch of $m$ examples with inputs $X$ and targets $Y$, let $Z = (z_1, \dots, z_m) \in \real^{d \times m}$ be the embeddings of the mini-batch, where $Z = f_\theta(X)$. \emph{Manifold mixup}~\cite{verma2019manifold} interpolates the embeddings and targets by forming a convex combination of the pairs with interpolation factor $\lambda \in [0,1]$:
\begin{align}
	\wt{Z} &= Z (\lambda I + (1 - \lambda) \Pi) \label{eq:embed} \\
	\wt{Y} &= Y (\lambda I + (1 - \lambda) \Pi) \label{eq:target},
\end{align}
where $\lambda \sim \Beta(\alpha, \alpha)$,
$I$ is the identity matrix
and $\Pi \in \real^{m \times m}$ is a permutation matrix. \emph{Input mixup}~\cite{zhang2018mixup} interpolates inputs rather than embeddings:
\begin{align}
	\wt{X} = X (\lambda I + (1 - \lambda) \Pi).
\label{eq:input}
\end{align}
Whatever the interpolation method and the space where it is performed, the interpolated data, \eg $\wt{X}$~\cite{zhang2018mixup} or $\wt{Z}$~\cite{verma2019manifold}, replaces the original mini-batch data and gives rise to predicted probabilities $\wt{P} = (p_1, \dots, p_m) \in \real^{c \times m}$ over classes, \eg $\wt{P} = f(\wt{X})$~\cite{zhang2018mixup} or $\wt{P} = g_W(\wt{Z})$~\cite{verma2019manifold}. Then, the average cross-entropy $H(\wt{Y}, \wt{P})$~\eq{ce} between the predicted probabilities $\wt{P}$ and interpolated targets $\wt{Y}$ is minimized.

The number of interpolated data is $m$, same as the original mini-batch data.


\subsection{MultiMix}
\label{sec:multi}

\paragraph{Interpolation}

Given a mini-batch of $m$ examples with embeddings $Z$ and targets $Y$, we draw interpolation vectors $\lambda_k \sim \Dir(\alpha)$ for $k = 1,\dots, n$, where $\Dir(\alpha)$ is the symmetric Dirichlet distribution
and $\lambda_k \in \Delta^{m-1}$, that is, $\lambda_k \ge 0$ and $\vone_m\tran \lambda_k = 1$. We then interpolate embeddings and targets by taking $n$ convex combinations over all $m$ examples:
\begin{align}
    \wt{Z} &= Z \Lambda \label{eq:multi-embed} \\
    \wt{Y} &= Y \Lambda, \label{eq:multi-target}
\end{align}
where $\Lambda = (\lambda_1, \dots, \lambda_n) \in \real^{m \times n}$. We thus generalize manifold mixup~\cite{verma2019manifold}:
\begin{enumerate}
	\item from pairs to tuples of length $m$, as long as the mini-batch: $m$-term convex combination~\eq{multi-embed},\eq{multi-target} \vs 2-term in~\eq{embed},\eq{target}, Dirichlet \vs Beta distribution;
	\item from $m$ to an arbitrary number $n$ of tuples: interpolated embeddings $\wt{Z} \in \real^{d \times n}$~\eq{multi-embed} \vs $\real^{d \times m}$ in~\eq{embed}, interpolated targets $\wt{Y} \in \real^{c \times n}$~\eq{multi-target} \vs $\real^{c \times m}$ in~\eq{target};
	\item from fixed $\lambda$ across the mini-batch to a different $\lambda_k$ for each interpolated item.
\end{enumerate}

\paragraph{Loss}

Again, we replace the original mini-batch embeddings $Z$ by the interpolated embeddings $\wt{Z}$ and minimize the average cross-entropy $H(\wt{Y}, \wt{P})$~\eq{ce} between the predicted probabilities $\wt{P} = g_W(\wt{Z})$ and the interpolated targets $\wt{Y}$~\eq{multi-target}. Compared with~\eq{loss}, the mini-batch loss becomes
\begin{align}
	L_M(X,Y;\theta,W) \defn H(Y \Lambda, g_W(f_\theta(X) \Lambda)).
\label{eq:loss-multi}
\end{align}


\subsection{MultiMix with self-distillation}
\label{sec:distill}

\paragraph{Networks}

We use an online self-distillation approach whereby the network $f \defn g_W \circ f_\theta$ that we learn becomes the \emph{student}, whereas a \emph{teacher} network $f' \defn g_{W'} \circ f_{\theta'}$ of the same architecture is obtained by exponential moving average of the parameters~\cite{tarvainen2017mean, grill2020bootstrap}.
The teacher parameters $(\theta', W')$ are not learned: We stop the gradient in the computation graph.

\paragraph{Views}

Given two transformations $T$ and $T'$, we generate two different augmented views $v = t(x)$ and $v' = t'(x)$ for each input $x$, where $t \sim T$ and $t' \sim T'$. Then, given a mini-batch of $m$ examples with inputs $X$ and targets $Y$, let $V = t(X), V' = t'(X) \in \real^{D \times m}$ be the mini-batch views corresponding to the two augmentations and $Z = f_\theta(V), Z' = f_{\theta'}(V') \in \real^{d \times m}$ the embeddings obtained by the student and teacher encoders respectively.

\paragraph{Interpolation}

We obtain the interpolated embeddings $\wt{Z}, \wt{Z}'$ from $Z, Z'$ by~\eq{multi-embed} and targets $\wt{Y}$ from $Y$ by~\eq{multi-target}, using the same $\Lambda$. The predicted class probabilities are given by $\wt{P} = g_W(\wt{Z})$ and $\wt{P}' = g_{W'}(\wt{Z}')$, again obtained by the student and teacher classifiers, respectively.

\paragraph{Loss}

We learn parameters $(\theta,W)$ by minimizing a classification and a self-distillation loss:
\begin{align}
	\gamma H(\wt{Y}, \wt{P}) + (1-\gamma) H(\wt{P}', \wt{P}),
\label{eq:distill-1}
\end{align}
where $\gamma \in [0,1]$. The former brings the probabilities $\wt{P}$ predicted by the student close to the targets $\wt{Y}$, as in~\eq{loss-multi}. The latter brings $\wt{P}$ close to the probabilities $\wt{P}'$ predicted by the teacher.

\begin{figure}
\centering
\input{tex/fig-dense}
\caption{\emph{Dense MultiMix} (\autoref{sec:dense}) for the special case $m = 2$ (two examples), $n = 1$ (one interpolated embedding), $r = 9$ (spatial resolution $3 \times 3$). The embeddings $\vz_1,\vz_2 \in \real^{d \times 9}$ of input images $x_1, x_2$ are extracted by encoder $f_\theta$. Attention maps $a_1, a_2 \in \real^9$ are extracted~\eq{attn}, multiplied element-wise with interpolation vectors $\lambda, (1-\lambda) \in \real^9$~\eq{fact-scale} and $\ell_1$-normalized per spatial position~\eq{fact-norm}. The resulting weights are used to form the interpolated embedding $\tilde{\vz} \in \real^{d \times 9}$ as a convex combination of $\vz_1,\vz_2$ per spatial position~\eq{dense-embed}. Targets are interpolated similarly~\eq{dense-target}.}
\label{fig:dense}
\end{figure}

\subsection{Dense MultiMix}
\label{sec:dense}

We now extend the previous methodology to the case where the embeddings are structured, \eg in matrices or tensors rather than vectors. This happens \eg with token \vs sentence embeddings in NLP and patch \vs image embeddings in vision. In practice, this works by removing spatial pooling and rather applying the loss function densely over all tokens/patches. The idea is illustrated in~\autoref{fig:dense}. For the sake of exposition, the formulation below uses sets of matrices grouped either by example or by spatial position. In practice, all operations are on tensors.

\paragraph{Preliminaries}

The encoder is now $f_\theta: \cX \to \real^{d \times r}$, mapping the input $x$ to an embedding $\vz = f_\theta(x) \in \real^{d \times r}$, where $d$ is the number of channels and $r$ is its spatial resolution---if there are more than one spatial dimensions, these are flattened.

Given a mini-batch of $m$ examples, we have again inputs $X = (x_1, \dots, x_m) \in \real^{D \times m}$ and targets $Y = (y_1, \dots, y_m) \in \real^{c \times m}$. Each embedding $\vz_i = f_\theta(x_i) = (z_i^1, \dots, z_i^r) \in \real^{d \times r}$ for $i = 1, \dots, m$ consists of features $z_i^j \in \real^d$ for spatial position $j = 1, \dots, r$. We group features by position in matrices $Z^1, \dots, Z^r$, where $Z^j = (z_1^j, \dots, z_m^j) \in \real^{d \times m}$ for $j = 1, \dots, r$.

\paragraph{Attention}

Each feature vector will inherit the target of the corresponding input example. However, we also attach a level of confidence according to an attention map. Given an embedding $\vz \in \real^{d \times r}$ with target $y \in \cY$ and a vector $u \in \real^d$, the attention map
\begin{align}
	a & = h(\vz\tran u) \in \real^r
\label{eq:attn}
\end{align}
measures the similarity of features of $\vz$ to $u$, where $h$ is a non-linearity, \eg softmax or ReLU followed by $\ell_1$ normalization. There are different ways to define vector $u$. For example, $u = \vz \vone_r / r$ by global average pooling (GAP) of $\vz$, or $u = W y$ assuming a linear classifier with $W \in \real^{d \times c}$, similar to class activation mapping (CAM)~\cite{zhou2016learning}. In the case of no attention, $a = \vone_r / r$ is uniform.

Given a mini-batch, let $a_i = (a_i^1, \dots, a_i^r) \in \real^r$ be the attention map of embedding $\vz_i$~\eq{attn}. We group attention by position in vectors $a^1, \dots, a^r$, where $a^j = (a_1^j, \dots, a_m^j) \in \real^m$ for $j = 1, \dots, r$.

\paragraph{Interpolation}

For each spatial position $j = 1, \dots, r$, we draw interpolation vectors $\lambda_k^j \sim \Dir(\alpha)$ for $k = 1,\dots, n$ and define $\Lambda^j = (\lambda_1^j, \dots, \lambda_n^j) \in \real^{m \times n}$. Because input examples are assumed to contribute according to the attention vector $a^j \in \real^m$, we scale the rows of $\Lambda^j$ accordingly and then we normalize its columns back to $\Delta^{m-1}$ so that they can define convex combinations:
\begin{align}
	M^j & = \diag(a^j) \Lambda^j \label{eq:fact-scale} \\
	\hat{M}^j & = M^j \diag(\vone_m\tran M^j)^{-1} \label{eq:fact-norm}
\end{align}
We then interpolate embeddings and targets by taking $n$ convex combinations over $m$ examples:
\begin{align}
    \wt{Z}^j &= Z^j \hat{M}^j \label{eq:dense-embed} \\
    \wt{Y}^j &= Y \hat{M}^j. \label{eq:dense-target}
\end{align}
This is similar to~\eq{multi-embed},\eq{multi-target}, but there is a different interpolated embedding matrix $\wt{Z}^j \in \real^{d \times n}$ as well as target matrix $\wt{Y}^j \in \real^{c \times n}$ per position, even though the original target matrix $Y$ is one.

\paragraph{Classifier}

The classifier is now $g_W: \real^{d \times r} \to \real^{c \times r}$, maintaining the same spatial resolution as the embedding and generating one vector of predicted probabilities per spatial position. This is done by removing average pooling or any down-sampling operation. The interpolated embeddings $\wt{Z}^1, \dots, \wt{Z}^r$~\eq{dense-embed} are grouped by example into $\wt{\vz}_1, \dots, \wt{\vz}_n \in \real^{d \times r}$, mapped by $g_W$ to predicted probabilities $\wt{\vp}_1, \dots, \wt{\vp}_n \in \real^{c \times r}$ and grouped again by position into $\wt{P}^1, \dots, \wt{P}^r \in \real^{c \times n}$.

In the simple case where the original classifier is linear, \ie $W \in \real^{d \times c}$, it is seen as $1 \times 1$ convolution and applied densely to each column (feature) of $\wt{Z}^j$ for $j = 1, \dots, r$.

\paragraph{Loss}

Finally, we learn parameters $\theta, W$ by minimizing the weighted cross-entropy $H(\wt{Y}^j, \wt{P}^j; s)$ of $\wt{P}^j$ relative to the interpolated targets $\wt{Y}^j$ again densely at each position $j$, where
\begin{align}
	H(Y, P; s) \defn - \vone_c\tran (Y \odot \log(P)) s / (\vone_n\tran s)
\label{eq:wce}
\end{align}
generalizes~\eq{ce} and the weight vector is defined as $s = \vone_m\tran M^j \in \real^n$. This is exactly the vector used to normalize the columns of $M^j$ in~\eq{fact-norm}. The motivation is that the columns of $M^j$ are the original interpolation vectors weighted by attention: A small $\ell_1$ norm indicates that for the given position $j$, we are sampling from examples of low attention, hence the loss is to be discounted.

\section{Experiments}
\label{sec:exp}

\subsection{Setup}
\label{sec:setup}

We use a mini-batch of size $m = 128$ examples in all experiments. For every mini-batch, we apply MultiMix with probability $0.5$ or input mixup otherwise. For MultiMix, the default settings are given in \autoref{sec:ablation}. We follow the experimental settings of AlignMixup~\citep{venkataramanan2021alignmix} and use PreActResnet-18 (R-18)~\citep{he2016deep} and WRN16-8~\citep{zagoruyko2016wide} as encoder on CIFAR-10 and CIFAR-100 datasets~\cite{krizhevsky2009learning}; R-18 on TinyImagenet~\citep{yao2015tiny} (TI); and Resnet-50 (R-50) and ViT-S/16~\citep{dosovitskiy2020image} on ImageNet~\citep{russakovsky2015imagenet}. We use top-1 error (\%) as evaluation metric on image classification and robustness to adversarial attacks (\autoref{sec:class}). We also experiment on object detection (\autoref{sec:det}) and out-of-distribution detection, which is in the supplementary material along with more details and results.

\subsection{Results: Image classification and robustness}
\label{sec:class}

\begin{table}
\begin{subtable}[T]{0.51\textwidth}
\centering
\scriptsize
\setlength{\tabcolsep}{4.5pt}
\begin{tabular}{lccccc} \toprule
	\Th{Dataset}                                            & \mc{2}{\Th{Cifar-10}}             & \mc{2}{\Th{Cifar-100}}            & TI              \\
	\Th{Network}                                            & R-18            & W16-8           & R-18            & W16-8           & R-18            \\ \midrule
	Baseline$^\dagger$                                      & 5.19            & 5.11            & 23.24           & 20.63           & 43.40           \\
	Manifold mixup~\citep{verma2019manifold}$^\dagger$      & 2.95            & 3.56            & 19.80           & 19.23           & 40.76           \\
	PuzzleMix~\citep{kim2020puzzle}$^\dagger$               & 2.93            & \ul{2.99}       & 20.01           & 19.25           & 36.52           \\
	Co-Mixup~\citep{kim2021co}$^\dagger$                    & \ul{2.89}       & 3.04            & 19.81           & 19.57           & 35.85           \\
	AlignMixup~\citep{venkataramanan2021alignmix}$^\dagger$ & 2.95            & 3.09            & \ul{18.29}      & \ul{18.77}      & \ul{33.13}      \\ \midrule
	MultiMix (ours)                                         & 2.97            & 2.92            & 18.19           & 18.57           & 32.78           \\
	\hspace{3pt}$+$ distil                                  & \se{2.87}       & \tb{2.78}       & \se{17.72}      & \se{17.91}      & 31.93           \\
	\hspace{3pt}$+$ dense                                   & 2.91            & \se{2.89}       & 18.12           & 18.20           & \se{31.54}           \\
	\rowcolor{LightCyan}
	\hspace{3pt}$+$ dense $+$ distil                        & \tb{2.81}       & \tb{2.78}       & \tb{17.48}      & \tb{17.66}      & \tb{30.87}      \\ \midrule
	Gain                                                    & \gp{\tb{+0.08}} & \gp{\tb{+0.21}} & \gp{\tb{+0.81}} & \gp{\tb{+1.11}} & \gp{\tb{+2.26}} \\ \bottomrule
\end{tabular}
\caption{\emph{Image classification} top-1 error (\%) on CIFAR-10/100 and TI (TinyImagenet). R: PreActResnet, W: WRN.}
\end{subtable}
\hspace{6pt}
\begin{subtable}[T]{0.45\textwidth}
\centering
\scriptsize
\setlength{\tabcolsep}{2pt}
\begin{tabular}{lcccc} \toprule
	\Th{Network}                                              & \mc{2}{\Th{Resnet-50}}            & \mc{2}{\Th{ViT-S/16}}              \\
	\Th{Method}                                               & \Th{Speed}      & \Th{Error}      & \Th{Speed}      & \Th{Error}       \\ \midrule
	Baseline$^\dagger$                                        & 1.17            & 23.68           & 1.01            & 26.1             \\
	Manifold mixup~\citep{verma2019manifold}$^\dagger$        & 1.15            & 22.50           & 0.97            & 24.8             \\
	PuzzleMix~\citep{kim2020puzzle}$^\dagger$                 & 0.84            & 21.24           & 0.73            & 24.3             \\
	Co-Mixup~\citep{kim2021co}$^\dagger$                      & 0.62            & --              & 0.57            & \ul{24.1}        \\
	AlignMixup~\citep{venkataramanan2021alignmix}$^\dagger$   & 1.03            & \ul{20.68}      & --              & --               \\ \midrule
	MultiMix (ours)                                           & 1.16            & 21.19           & 1.0             & 24.8             \\
	\hspace{3pt}$+$ distil                                    & 1.06            & \se{19.88}      & 0.93            & \se{23.4}        \\
	\hspace{3pt}$+$ dense                                     & 0.95            & 20.63           & 0.88            & 23.9             \\
	\rowcolor{LightCyan}
	\hspace{3pt}$+$ dense $+$ distil                          & 0.83            & \tb{19.79}      & 0.81            & \tb{23.1}        \\ \midrule
	Gain                                                      &                 & \gp{\tb{+0.89}} &                 & \gp{\tb{+1.0}}   \\ \bottomrule
\end{tabular}
\caption{\emph{Image classification and training speed} on ImageNet. Top-1 error (\%): lower is better. Speed: images/sec ($\times 10^3$): higher is better.}
\end{subtable}
\caption{\emph{Image classification and training speed.} $^\dagger$: reported by AlignMixup. \tb{Bold black}: best; \se{Blue}: second best; underline: best baseline. Gain: reduction of error over best baseline. Comparison with additional baselines is given in the supplementary material.}
\label{tab:cls}
\end{table}

\paragraph{Image classification}

In \autoref{tab:cls}(a) we observe that MultiMix and Dense MultiMix already outperform SoTA on all datasets except CIFAR-10 with R-18, where they are on par with Co-Mixup. The addition of distillation increases the gain and outperforms SoTA on all datasets. Both distillation and dense improve over vanilla MultiMix and their effect is complementary on all datasets. On TI for example, distillation improves by 0.85\%, dense by 1.24\% and their combination by 1.91\%. This combination brings an impressive gain of 2.26\% over the previous SoTA -- AlignMixup.

In \autoref{tab:cls}(b) we observe that on ImageNet with R-50, vanilla MultiMix already outperforms all methods except SoTA AlignMixup. The addition of dense, distillation or both outperforms all SoTA with both R-50 and ViT-S/16. More importantly, it brings an overall gain of 4\% over the baseline with R-50 and 3\% with ViT-S/16.

\paragraph{Training speed}

\autoref{tab:cls}(b) shows the training speed of MultiMix and its variants compared with SoTA mixup methods, measured on NVIDIA V-100 GPU, including forward and backward pass. In terms of training speed, the vanilla MultiMix is on par with the baseline, bringing a gain of 2.49\%. The addition of distillation is on par with SoTA AlignMixup, bringing a gain of 0.80\%. Adding both dense and distillation brings a gain of 0.89\% over AlignMixup, while being 19.4\% slower. The inference speed is the same for all methods.


\paragraph{Robustness to adversarial attacks}

We follow the experimental settings of AlignMixup~\citep{venkataramanan2021alignmix} and use $8/255$ $ l_\infty$ $\epsilon$-ball for FGSM~\citep{goodfellow2015explaining} and $4/255$ $ l_\infty$ $\epsilon$-ball with step size 2/255 for PGD~\citep{madry2017towards} attack. In \autoref{tab:adv} we observe that vanilla MultiMix is already more robust than SoTA on all datasets and settings except FGSM on CIFAR-100 with R-18, where it is on par with AlignMixup. The addition of dense, distillation or both again increases the robustness and shows that their effect is complementary. The overall gain is more impressive than in classification error. For example, against the strong PGD attack on CIFAR-10 with W16-8, the SoTA Co-Mixup improves the baseline by 3.75\% and our best result improves the baseline by 9.38\%, which is more than double.

\begin{wraptable}{r}{6.2cm}
\vspace{-12pt}
\centering
\scriptsize
\setlength{\tabcolsep}{3pt}
\begin{tabular}{lcccc} \toprule
	\Th{Dataset}                     & \Th{VOC07$+$12}        & \Th{MS-COCO}         \\
	\Th{Detector}                    & \Th{SSD}               & \Th{Faster R-CNN}    \\ \midrule
	Baseline$^\dagger$               & 76.7                   & 33.27                \\
	Input mixup$^\dagger$            & 76.6                   & 34.18                \\
	CutMix$^\dagger$                 & 77.6                   & 35.16                \\
	AlignMixup$^\dagger$             & \ul{78.4}              & \ul{35.84}           \\ \midrule
	MultiMix (ours)                  & 77.9                   & 35.73                \\
	\hspace{3pt}$+$ distil           & \se{78.7}              & \se{35.97}           \\
	\hspace{3pt}$+$ dense            & 78.5                   & 35.89                \\
	\rowcolor{LightCyan}
	\hspace{3pt}$+$ dense $+$ distil & \tb{79.1}              & \tb{36.41}           \\ \midrule
	Gain                             & \gp{\tb{+0.7}}         & \gp{\tb{+0.57}}      \\ \bottomrule
\end{tabular}
\caption{\emph{Transfer learning} to object detection. Mean average precision (mAP, \%): higher is better. $^\dagger$: reported by AlignMixup. \tb{Bold black}: best; \se{Blue}: second best; underline: best baseline. Gain: increase in mAP.}
\label{tab:obj-det}
\vspace{-12pt}
\end{wraptable}

\subsection{Results: Transfer learning to object detection}
\label{sec:det}

We evaluate the effect of mixup on the generalization ability of a pre-trained network to object detection as a downstream task. Following the settings of CutMix~\citep{yun2019cutmix}, we pre-train R-50 on ImageNet with MultiMix and its variants and use it as the backbone for SSD~\citep{liu2016ssd} with fine-tuning on Pascal VOC07$+$12~\citep{everingham2010pascal} and Faster-RCNN~\citep{ren2015faster} with fine-tuning on MS-COCO~\citep{lin2014microsoft}.

In \autoref{tab:obj-det}, we observe that, while vanilla MultiMix is slightly worse than SoTA AlignMixup, dense and distillation bring improvements over the SoTA on both datasets and are still complementary. This is consistent with classification results. Compared with the baseline, our best setting brings a gain of 2.40\% mAP on Pascal VOC07$+$12 and 3.14\% on MS-COCO.

\begin{table}[t!]
\centering
\scriptsize
\setlength{\tabcolsep}{6pt}
\begin{tabular}{lccccc|cccc} \toprule
	\Th{Attack}                                             & \multicolumn{5}{c}{FGSM}                                                                & \multicolumn{4}{c}{PGD}                                               \\ \midrule
	\Th{Dataset}                                            & \mc{2}{\Th{Cifar-10}}             & \mc{2}{\Th{Cifar-100}}            & TI              & \mc{2}{\Th{Cifar-10}}             & \mc{2}{\Th{Cifar-100}}            \\
	\Th{Network}                                            & R-18            & W16-8           & R-18            & W16-8           & R-18            & R-18            & W16-8           & R-18            & W16-8           \\ \midrule
	Baseline$^\dagger$                                      & 89.41           & 88.02           & 87.12           & 72.81           & 91.85           & 99.99           & 99.94           & 99.97           & 99.99           \\
	Manifold mixup~\citep{verma2019manifold}$^\dagger$      & 77.63           & 76.11           & 80.29           & 56.45           & 89.25           & 97.22           & 98.49           & 99.66           & 98.43           \\
	PuzzleMix~\citep{kim2020puzzle}$^\dagger$               & 57.11           & 60.73           & 78.70           & 57.77           & 83.91           & 97.73           & 97.00           & 96.42           & 95.28           \\
	Co-Mixup~\citep{kim2021co}$^\dagger$                    & 60.19           & 58.93           & 77.61           & 56.59           & --              & 97.59           & \ul{96.19}      & 95.35           & 94.23           \\
	AlignMixup~\citep{venkataramanan2021alignmix}$^\dagger$ & \ul{54.83}      & \ul{56.20}      & \ul{74.18}      & \ul{55.05}      & \ul{78.83}      & \ul{95.42}      & 96.71           & \ul{90.40}      & \ul{92.16}      \\ \midrule
	MultiMix (ours)                                         & 54.19           & 55.39           & 75.84           & 54.58           & 77.51           & 94.27           & 94.83           & 90.02           & 91.68           \\
	\hspace{3pt}$+$ distillation                            & \se{52.55}      & \se{51.42}      & \se{73.55}      & \se{52.77}      & 76.20           & \se{92.69}      & 93.90           & 88.87           & \se{90.54}      \\
	\hspace{3pt}$+$ dense                                   & 54.10           & 53.33           & 74.48           & 53.01           & \se{75.57}      & 92.99           & \se{92.68}      & \se{88.60}      & 90.90           \\
	\rowcolor{LightCyan}
	\hspace{3pt}$+$ dense $+$ distillation                  & \tb{52.07}      & \tb{50.17}      & \tb{72.98}      & \tb{52.19}      & \tb{75.18}      & \tb{90.82}      & \tb{90.56}      & \tb{87.58}      & \tb{90.18}      \\ \midrule
	Gain                                                    & \gp{\tb{+2.76}} & \gp{\tb{+5.95}} & \gp{\tb{+1.20}} & \gp{\tb{+2.86}} & \gp{\tb{+3.65}} & \gp{\tb{+4.60}} & \gp{\tb{+5.63}} & \gp{\tb{+2.82}} & \gp{\tb{+1.98}}  \\ \bottomrule
\end{tabular}
\vspace{6pt}
\caption{\emph{Robustness to FGSM \& PGD attacks}. Top-1 error (\%): lower is better. $^\dagger$: reported by AlignMixup. \tb{Bold black}: best; \se{Blue}: second best; underline: best baseline. Gain: reduction of error over best baseline. TI: TinyImagenet. R: PreActResnet, W: WRN.  Comparison
with additional baselines is given in the supplementary material.}
\label{tab:adv}
\end{table}


\subsection{Analysis of the embedding space}

\begin{figure}
\centering
\scriptsize
\setlength{\tabcolsep}{2.5pt}
\newcommand{\sz}{.19}

\vspace{-10pt}
\begin{tabular}{ccccc}
	\fig[\sz]{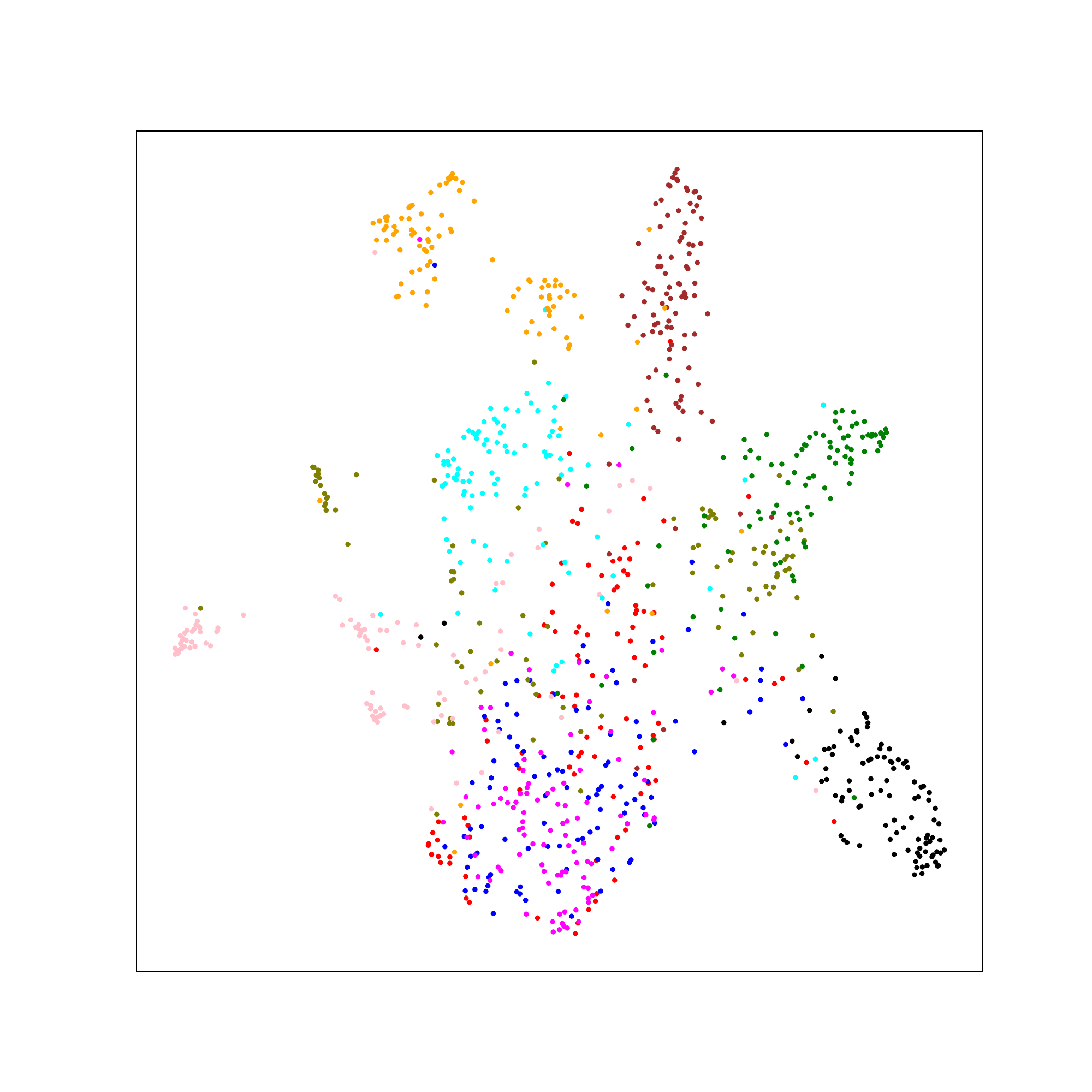} &
	\fig[\sz]{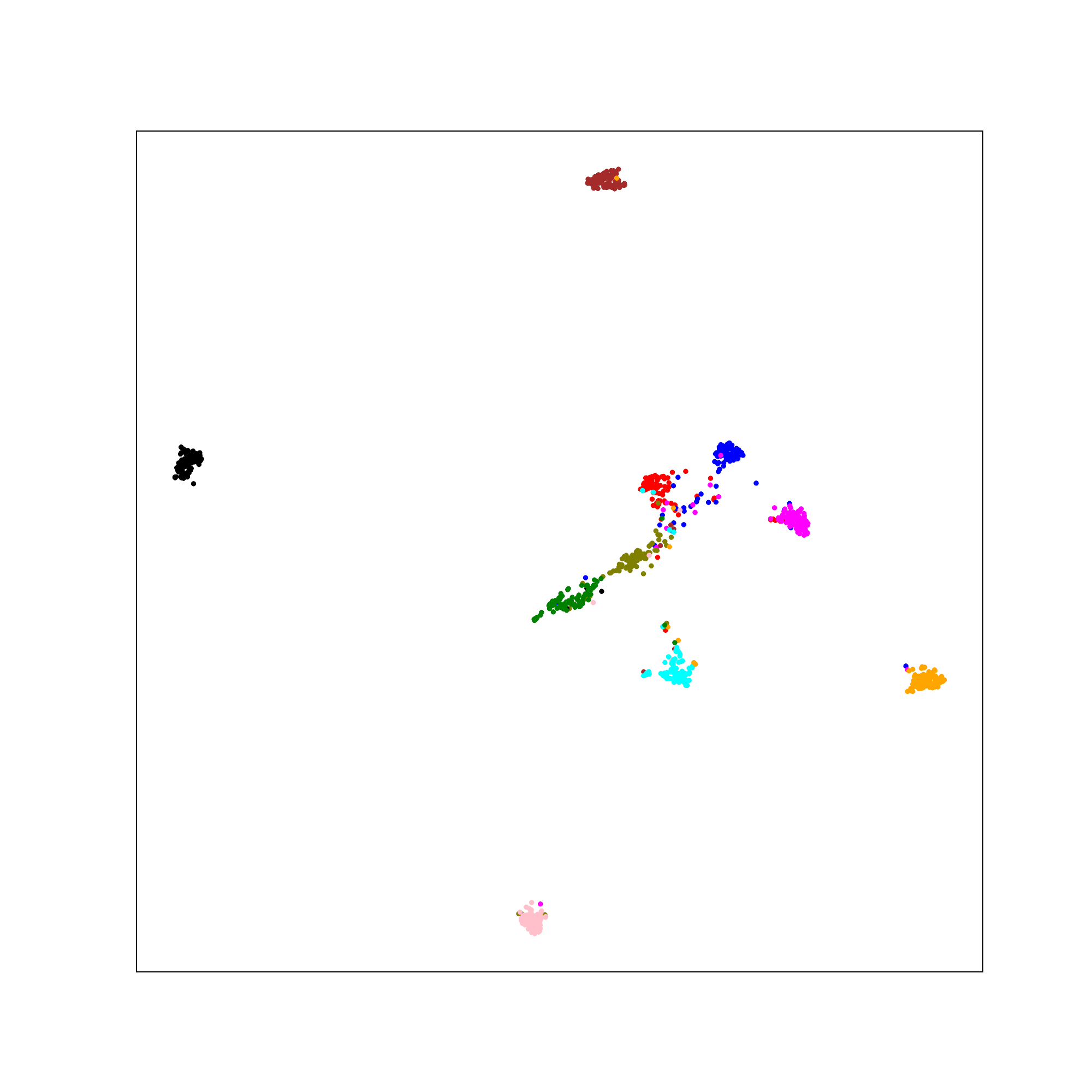} &
	\fig[\sz]{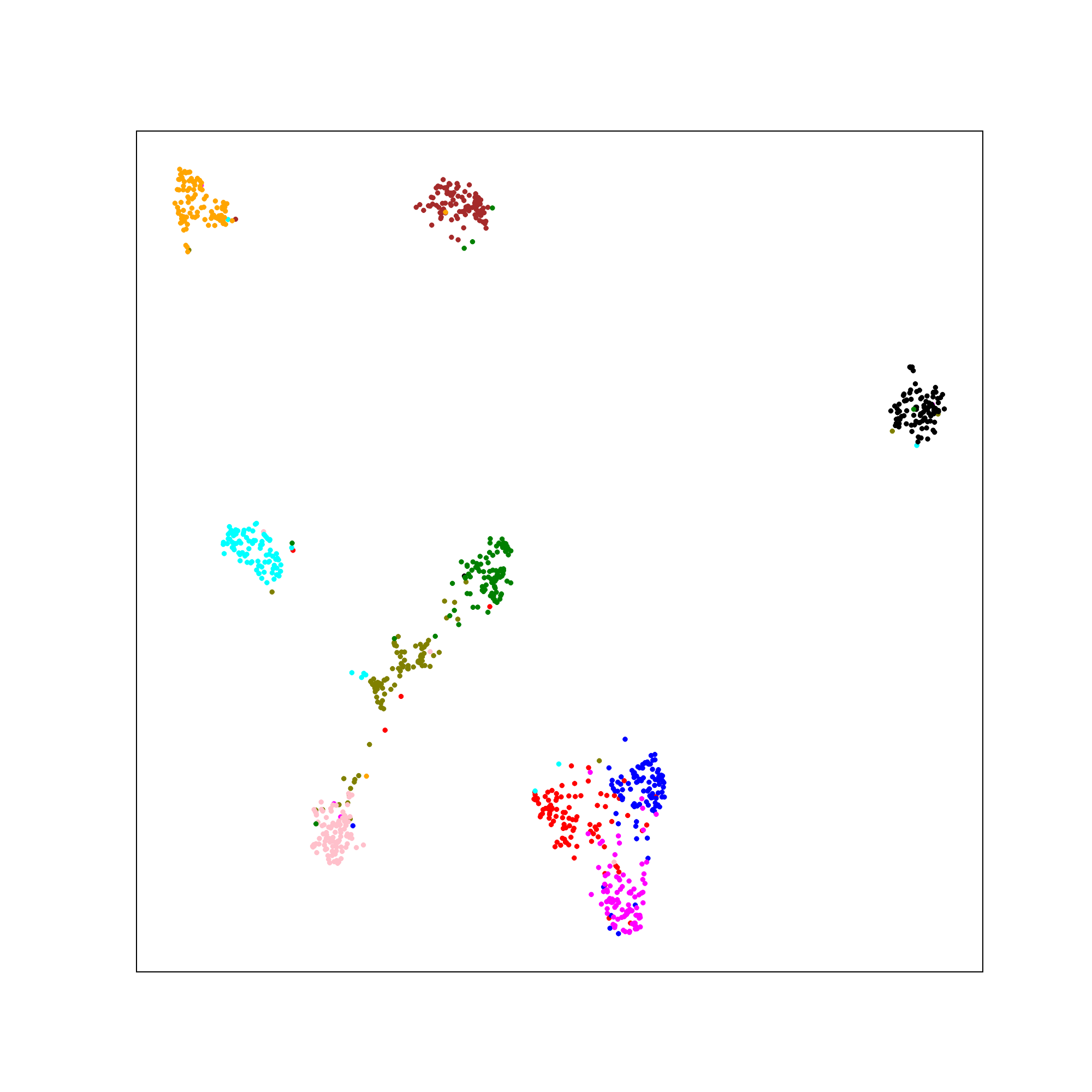}  &
	\fig[\sz]{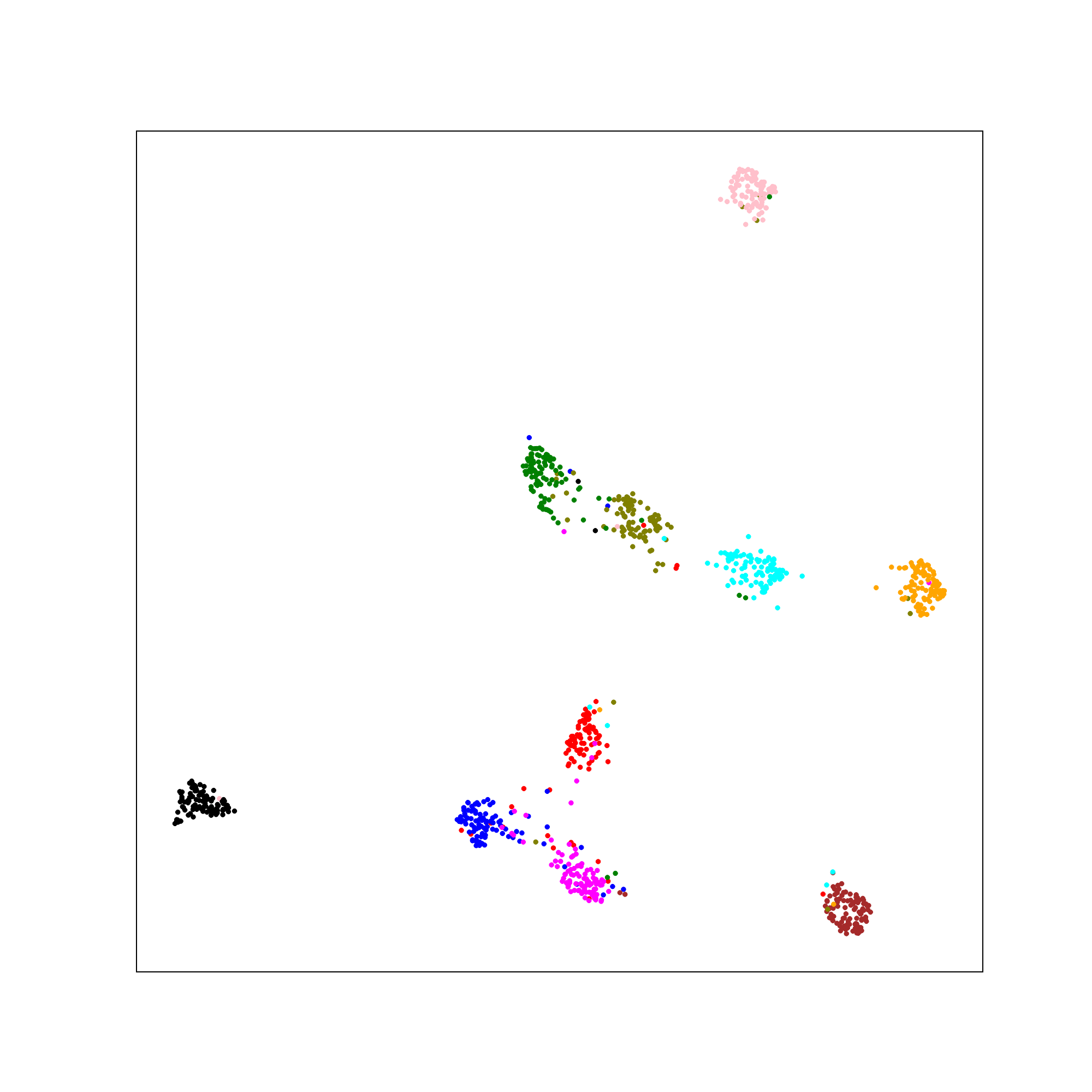} &
	\fig[\sz]{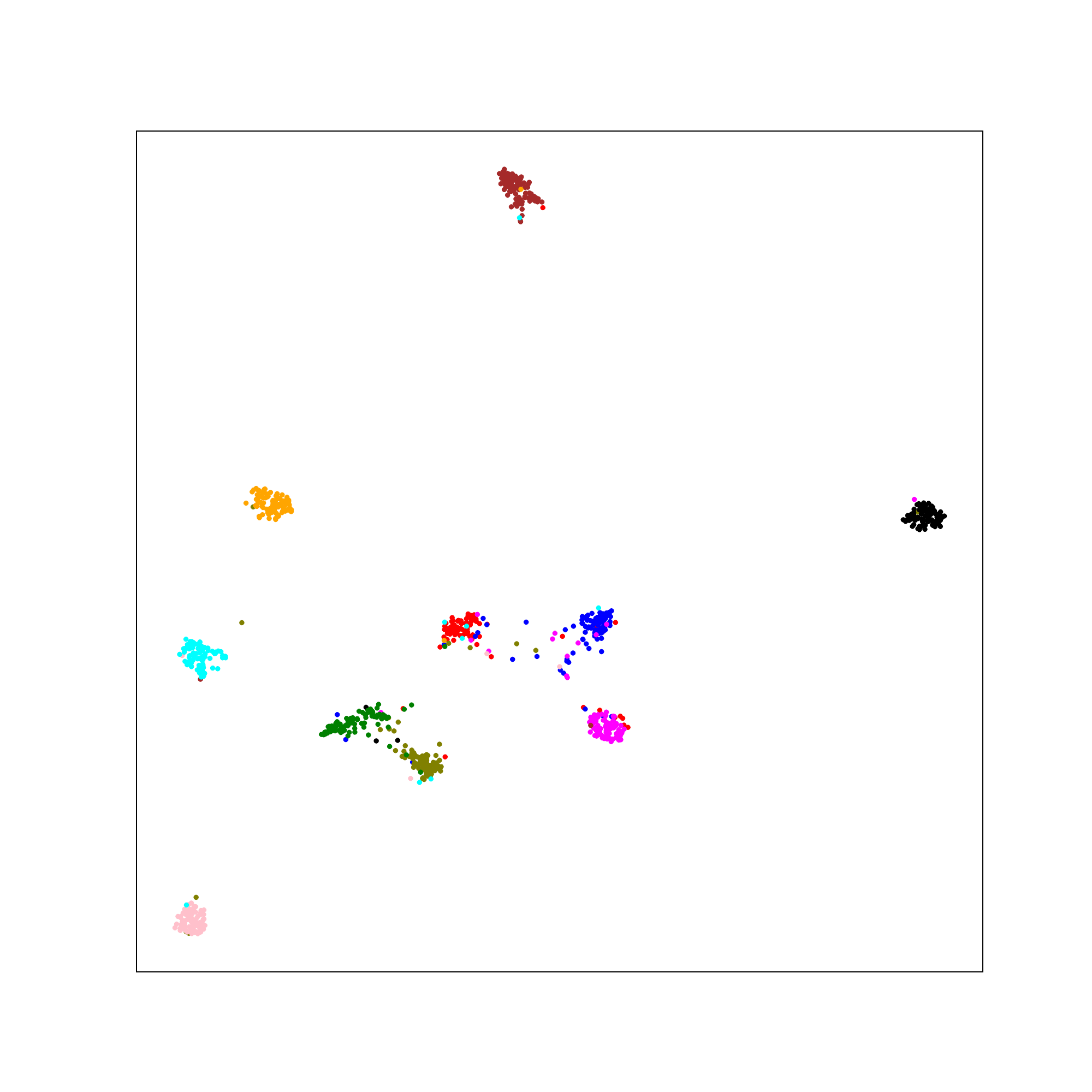} \\[3pt]
	(a) Baseline &
	(b) Manifold mixup~\citep{verma2019manifold} &
	(c) SaliencyMix~\citep{uddin2020saliencymix} &
	(d) AlignMixup~\citep{venkataramanan2021alignmix} &
	(e) Dense MultiMix $+$ \\
	               &
	               &
	               &
	               &
	distillation (ours)
\end{tabular}
\caption{\emph{Embedding space visualization} for 100 test examples per class of 10 randomly chosen classes of CIFAR-100 with PreActResnet-18, using UMAP~\citep{mcinnes2018umap}.}
\vspace{-10pt}
\label{fig:embedding_space}
\end{figure}

\paragraph{Qualitative analysis}

We qualitatively analyze the embedding space on 10 CIFAR-100 classes in \autoref{fig:embedding_space}. We observe that the quality of embeddings of the baseline is extremely poor with severely overlapping classes, which explains its poor performance on image classification. All mixup methods result in clearly better clustered and more uniformly spread classes. Manifold mixup~\citep{verma2019manifold} produces five tightly clustered classes but the other five are still severely overlapping. SaliencyMix~\citep{uddin2020saliencymix} and AlignMixup~\citep{venkataramanan2021alignmix} yield four somewhat clustered classes and 6 moderately overlapping ones. Our best setting, \ie, dense MultiMix with distillation, results in five tightly clustered classes and another five somewhat overlapping but less than all competitors. More plots including variants of MultiMix are given in the supplementary material.

\paragraph{Quantitative analysis}

We also quantitatively assess the embedding space on the CIFAR-100 test set using alignment and uniformity~\citep{wang2020understanding}. \emph{Alignment} measures the expected pairwise distance of examples in the same class. Lower alignment indicates that the classes are more tightly clustered. \emph{Uniformity} measures the (log of the) expected pairwise similarity of all examples using a Gaussian kernel as a similarity function. Lower uniformity indicates that classes are more uniformly spread in the embedding space.
On CIFAR-100, we obtain alignment 3.02 for baseline, 1.27 for Manifold Mixup~\citep{verma2019manifold}, 2.44 for SaliencyMix~\citep{uddin2020saliencymix}, 2.04 for AlignMixup and 0.92 for Dense MultiMix with distillation. We also obtain uniformity -1.94 for the baseline, -2.38 for Manifold Mixup~\citep{verma2019manifold}, -2.82 for SaliencyMix~\citep{uddin2020saliencymix}, -4.77 for AlignMixup~\citep{venkataramanan2021alignmix} and -5.68 for dense MultiMix with distillation. These results validate the qualitative analysis of \autoref{fig:embedding_space}.
\subsection{Ablations}
\label{sec:ablation}

\begin{figure}
\centering
\setlength{\tabcolsep}{3pt}
{\scriptsize\ref*{named_cifar}}
\begin{tabular}{ccc}
\begin{tikzpicture}
\begin{axis}[
	width=5.1cm,
	height=4cm,
	font=\scriptsize,
	xlabel={(a) Mixing layers},
	ylabel={Accuracy},
	enlarge x limits=false,
	ymin=78.0, ymax=83,
	xtick={0,...,4},
	xticklabels={{0},{0,1},{0,2},{0,3},{0,4}},
]
	\pgfplotstableread{
		x    mm     mm-d   dmm    dmm-d
		0    79.79  79.79  79.79  79.79
		1    78.48  79.80  79.62  80.17
		2    79.18  79.37  80.92  81.43
		3    80.65  81.06  81.17  81.55
		4    81.81  82.28  81.88  82.58
	}{\lay}
	\addplot[blue,mark=*] table[y=mm]        {\lay}; 
	\addplot[blue,mark=*,dashed] table[y=mm-d] {\lay}; 
	\addplot[red,mark=*] table[y=dmm]        {\lay}; 
	\addplot[red,mark=*,dashed] table[y=dmm-d] {\lay}; 
\end{axis}
\end{tikzpicture}
&
\begin{tikzpicture}
\begin{axis}[
	width=5.1cm,
	height=4cm,
	font=\scriptsize,
	xlabel={(b) \# tuples $n$},
	enlarge x limits=false,
	ymin=80.0, ymax=83,
	xtick={0,...,4},
	xticklabels={$10^1$,$10^2$,$10^3$, $10^4$, $10^5$},
	legend columns=4,
	legend to name=named_cifar,
]
	\pgfplotstableread{
		x    mm     mm-d   dmm    dmm-d
		0    81.34  81.78  81.41  81.65
		1    81.73  82.23  81.61  82.13
		2    81.81  82.28  81.88  82.58
		3    81.83  82.39  81.89  82.50
		4    81.91  82.38  81.84  82.49
	}{\tup}
	\addplot[blue,mark=*] table[y=mm]        {\tup}; \leg{MultiMix}
	\addplot[blue,mark=*,dashed] table[y=mm-d] {\tup}; \leg{MultiMix $+$ distill}
	\addplot[red,mark=*] table[y=dmm]        {\tup}; \leg{Dense MultiMix}
	\addplot[red,mark=*,dashed] table[y=dmm-d] {\tup}; \leg{Dense MultiMix $+$ distill}
\end{axis}
\end{tikzpicture}
&
\begin{tikzpicture}
\begin{axis}[
	width=5.1cm,
	height=4cm,
	font=\scriptsize,
	xlabel={(c) Dirichlet parameter $\alpha$},
	enlarge x limits=false,
	ymin=80.0, ymax=83,
]
	\pgfplotstableread{
		x    mm     mm-d   dmm    dmm-d
		0.5    81.09  81.73  81.43  81.48
		1.0    81.21  82.12  81.59  82.23
		1.5    80.90  81.77  81.61  81.99
		2.0    80.33  81.58  81.62  81.80
	}{\alph}
	\addplot[blue,mark=*] table[y=mm]        {\alph}; 
	\addplot[blue,mark=*,dashed] table[y=mm-d] {\alph}; 
	\addplot[red,mark=*] table[y=dmm]        {\alph}; 
	\addplot[red,mark=*,dashed] table[y=dmm-d] {\alph}; 
\end{axis}
\end{tikzpicture}
\end{tabular}
\caption{\emph{Ablation study} of MultiMix and its variants on CIFAR-100 using R-18. (a) Interpolation layers (R-18 block; 0: input mixup). (b) Number of tuples $n$. (c) Dirichlet parameter $\alpha$.}
\vspace{-6pt}
\label{fig:ablation-plot}
\end{figure}
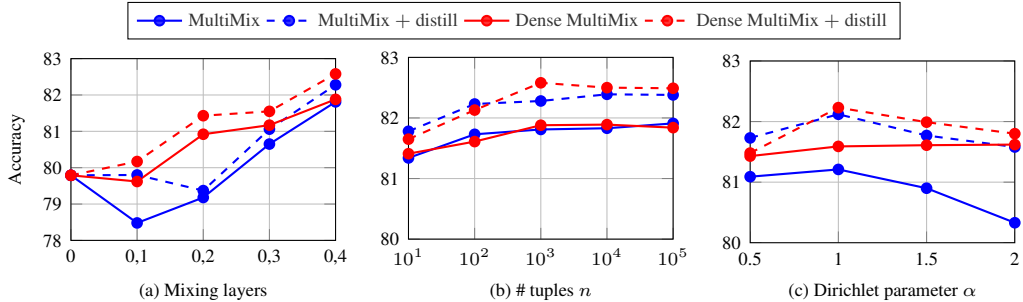

All ablations are performed using R-18 on CIFAR-100. For MultiMix, we study the effect of the layer where we interpolate, the number of tuples $n$ and a fixed value of Dirichlet parameter $\alpha$.

More ablations are given in the supplementary material.

\paragraph{Interpolation layer}

For MultiMix, we use the entire network as the encoder $f_\theta$ by default, except for the last fully-connected layer, which we use as classifier $g_W$. Thus, we interpolate embeddings in the deepest layer by default. Here, we study the effect of different decompositions of the network $f = g_W \circ f_\theta$, such that interpolation of embeddings takes place at a different layer. When using distillation, we interpolate at the same layer for both the teacher and the student. In \autoref{fig:ablation-plot}(a), we observe that mixing at the deeper layers of the network significantly improves performance. The same behavior is observed when adding dense, distillation, or both. This validates our default choice.

\paragraph{Number of tuples $n$}

Since our aim is to increase the amount of data seen by the model, or at least part of the model, it is important to study the number $n$ of interpolated embeddings. We observe from \autoref{fig:ablation-plot}(b) that accuracy increases overall with $n$ and saturates for $n \ge 1000$ for all variants of MultiMix. Our best setting, Dense MultiMix with distillation, works best at $n = 1000$. We choose this as default, given also that the training cost increases with $n$. The training speed as a function of $n$ is given in the supplementary material and is nearly constant for $n \le 1000$.

\paragraph{Dirichlet parameter $\alpha$}

Our default setting for $\alpha$ is to draw it uniformly at random from the interval $[0.5, 2]$ for every interpolation vector (column of $\Lambda$) that we draw. Here we study the effect of a fixed value of $\alpha$. In \autoref{fig:ablation-plot}(c), we observe that the best accuracy is achieved for $\alpha=1$ for most variants of MultiMix, which corresponds to the uniform distribution over the convex hull of the mini-batch embeddings. However, all measurements are lower than the default $\alpha \sim U[0.5, 2]$. For example, from \autoref{tab:cls}(a) (CIFAR-100, R-18), dense MultiMix + distillation has accuracy 82.52, compared with 82.23 in \autoref{fig:ablation-plot}(c) for $\alpha = 1$.

\section{Conclusion}
\label{sec:conclusion}

In terms of input interpolation, the take-home message of this work is that, instead of devising smarter and more complex interpolation functions in the input space or the first layers of the representation, it is more beneficial to just perform linear interpolation in the very last layer where the cost is minimal, and then increase as much as possible the number of interpolated embeddings for mixup. This is more in line with the original motivation of mixup as a way to go beyond ERM. In terms of target interpolation, the take-home message is the opposite: instead of linear interpolation of original targets, find new synthetic targets for the interpolated embeddings with the help of the network itself, then interpolate them linearly. This idea fits nicely with self-distillation, which is popular in settings such as self-supervised representation learning and continual learning. Interestingly, self-distillation can be seen as yet another form of augmentation, but in the model space.

A natural extension of this work is the application to settings other than supervised classification, which has been the focus of this work. A limitation is that it is not straightforward to combine the sampling scheme of MultiMix with complex interpolation methods, unless they are fast to compute in the embedding space.

\clearpage

\bibliography{egbib}

\begin{thebibliography}{10}\itemsep=-1pt

\bibitem{anil2018large}
Rohan Anil, Gabriel Pereyra, Alexandre Passos, Robert Ormandi, George~E Dahl,
  and Geoffrey~E Hinton.
\newblock Large scale distributed neural network training through online
  distillation.
\newblock {\em arXiv preprint arXiv:1804.03235}, 2018.

\bibitem{bucilua2006model}
Cristian Buciluǎ, Rich Caruana, and Alexandru Niculescu-Mizil.
\newblock Model compression.
\newblock In {\em ACM SIGKDD}, 2006.

\bibitem{caron2021emerging}
Mathilde Caron, Hugo Touvron, Ishan Misra, Herv{\'e} J{\'e}gou, Julien Mairal,
  Piotr Bojanowski, and Armand Joulin.
\newblock Emerging properties in self-supervised vision transformers.
\newblock In {\em ICCV}, 2021.

\bibitem{Dabouei_2021_CVPR}
Ali Dabouei, Sobhan Soleymani, Fariborz Taherkhani, and Nasser~M. Nasrabadi.
\newblock Supermix: Supervising the mixing data augmentation.
\newblock In {\em CVPR}, 2021.

\bibitem{bert}
Jacob Devlin, Ming-Wei Chang, Kenton Lee, and Kristina Toutanova.
\newblock Bert: Pre-training of deep bidirectional transformers for language
  understanding.
\newblock In {\em Proceedings of the 2019 Conference of the North American
  Chapter of the Association for Computational Linguistics: Human Language
  Technologies}, 2019.

\bibitem{Dhar_2019_CVPR}
Prithviraj Dhar, Rajat~Vikram Singh, Kuan-Chuan Peng, Ziyan Wu, and Rama
  Chellappa.
\newblock Learning without memorizing.
\newblock In {\em CVPR}, 2019.

\bibitem{dosovitskiy2020image}
Alexey Dosovitskiy, Lucas Beyer, Alexander Kolesnikov, Dirk Weissenborn,
  Xiaohua Zhai, Thomas Unterthiner, Mostafa Dehghani, Matthias Minderer, Georg
  Heigold, Sylvain Gelly, et~al.
\newblock An image is worth 16x16 words: Transformers for image recognition at
  scale.
\newblock {\em arXiv preprint arXiv:2010.11929}, 2020.

\bibitem{DCO+20}
Arthur Douillard, Matthieu Cord, Charles Ollion, Thomas Robert, and Eduardo
  Valle.
\newblock {PODNet}: Pooled outputs distillation for small-tasks incremental
  learning.
\newblock In {\em ECCV}, 2020.

\bibitem{everingham2010pascal}
Mark Everingham, Luc Van~Gool, Christopher~KI Williams, John Winn, and Andrew
  Zisserman.
\newblock The pascal visual object classes (voc) challenge.
\newblock {\em IJCV}, 2010.

\bibitem{goodfellow2015explaining}
Ian~J Goodfellow, Jonathon Shlens, and Christian Szegedy.
\newblock Explaining and harnessing adversarial examples.
\newblock In {\em ICLR}, 2015.

\bibitem{grill2020bootstrap}
Jean-Bastien Grill, Florian Strub, Florent Altch{\'e}, Corentin Tallec, Pierre
  Richemond, Elena Buchatskaya, Carl Doersch, Bernardo Avila~Pires, Zhaohan
  Guo, Mohammad Gheshlaghi~Azar, et~al.
\newblock Bootstrap your own latent-a new approach to self-supervised learning.
\newblock {\em NeurIPS}, 2020.

\bibitem{guo2019mixup}
Hongyu Guo, Yongyi Mao, and Richong Zhang.
\newblock Mixup as locally linear out-of-manifold regularization.
\newblock In {\em AAAI}, 2019.

\bibitem{he2017mask}
Kaiming He, Georgia Gkioxari, Piotr Doll{\'a}r, and Ross Girshick.
\newblock Mask {R-CNN}.
\newblock In {\em ICCV}, 2017.

\bibitem{he2016deep}
Kaiming He, Xiangyu Zhang, Shaoqing Ren, and Jian Sun.
\newblock Deep residual learning for image recognition.
\newblock In {\em CVPR}, 2016.

\bibitem{hinton2015distilling}
Geoffrey Hinton, Oriol Vinyals, Jeff Dean, et~al.
\newblock Distilling the knowledge in a neural network.
\newblock {\em arXiv preprint arXiv:1503.02531}, 2015.

\bibitem{hong2021stylemix}
Minui Hong, Jinwoo Choi, and Gunhee Kim.
\newblock Stylemix: Separating content and style for enhanced data
  augmentation.
\newblock In {\em CVPR}, 2021.

\bibitem{kim2021co}
Jang-Hyun Kim, Wonho Choo, Hosan Jeong, and Hyun~Oh Song.
\newblock Co-mixup: Saliency guided joint mixup with supermodular diversity.
\newblock In {\em ICLR}, 2021.

\bibitem{kim2020puzzle}
Jang-Hyun Kim, Wonho Choo, and Hyun~Oh Song.
\newblock Puzzle mix: Exploiting saliency and local statistics for optimal
  mixup.
\newblock In {\em ICML}, 2020.

\bibitem{krizhevsky2009learning}
Alex Krizhevsky, Geoffrey Hinton, et~al.
\newblock Learning multiple layers of features from tiny images.
\newblock Technical report, University of Toronto, 2009.

\bibitem{Li_2019_CVPR}
Wenbin Li, Lei Wang, Jinglin Xu, Jing Huo, Yang Gao, and Jiebo Luo.
\newblock Revisiting local descriptor based image-to-class measure for few-shot
  learning.
\newblock In {\em CVPR}, 2019.

\bibitem{li2021mst}
Zhaowen Li, Zhiyang Chen, Fan Yang, Wei Li, Yousong Zhu, Chaoyang Zhao, Rui
  Deng, Liwei Wu, Rui Zhao, Ming Tang, et~al.
\newblock {MST}: Masked self-supervised transformer for visual representation.
\newblock In {\em NeurIPS}, 2021.

\bibitem{LiHo18}
Z. Li and D. Hoiem.
\newblock Learning without forgetting.
\newblock {\em IEEE Transactions on Pattern Analysis and Machine Intelligence},
  40(12):2935--2947, Dec 2018.

\bibitem{Lifchitz_2019_CVPR}
Yann Lifchitz, Yannis Avrithis, Sylvaine Picard, and Andrei Bursuc.
\newblock Dense classification and implanting for few-shot learning.
\newblock In {\em CVPR}, 2019.

\bibitem{lin2014microsoft}
Tsung-Yi Lin, Michael Maire, Serge Belongie, James Hays, Pietro Perona, Deva
  Ramanan, Piotr Doll{\'a}r, and C~Lawrence Zitnick.
\newblock Microsoft coco: Common objects in context.
\newblock In {\em ECCV}, 2014.

\bibitem{liu2016ssd}
Wei Liu, Dragomir Anguelov, Dumitru Erhan, Christian Szegedy, Scott Reed,
  Cheng-Yang Fu, and Alexander~C Berg.
\newblock Ssd: Single shot multibox detector.
\newblock In {\em ECCV}, 2016.

\bibitem{madry2017towards}
Aleksander Madry, Aleksandar Makelov, Ludwig Schmidt, Dimitris Tsipras, and
  Adrian Vladu.
\newblock Towards deep learning models resistant to adversarial attacks.
\newblock In {\em ICLR}, 2018.

\bibitem{mcinnes2018umap}
Leland McInnes, John Healy, Nathaniel Saul, and Lukas Grossberger.
\newblock Umap: Uniform manifold approximation and projection.
\newblock {\em The Journal of Open Source Software}, 2018.

\bibitem{noh2015learning}
Hyeonwoo Noh, Seunghoon Hong, and Bohyung Han.
\newblock Learning deconvolution network for semantic segmentation.
\newblock In {\em ICCV}, 2015.

\bibitem{o2020unsupervised}
Pedro O~Pinheiro, Amjad Almahairi, Ryan Benmalek, Florian Golemo, and Aaron
  Courville.
\newblock Unsupervised learning of dense visual representations.
\newblock In {\em NeurIPS}, 2020.

\bibitem{RDG+18}
Ilija Radosavovic, Piotr Dollar, Ross Girshick, Georgia Gkioxari, and Kaiming
  He.
\newblock Data distillation: Towards omni-supervised learning.
\newblock In {\em CVPR}, 2018.

\bibitem{rebuffi2017icarl}
Sylvestre-Alvise Rebuffi, Alexander Kolesnikov, Georg Sperl, and Christoph~H
  Lampert.
\newblock {iCaRL}: Incremental classifier and representation learning.
\newblock In {\em CVPR}, 2017.

\bibitem{ren2015faster}
Shaoqing Ren, Kaiming He, Ross Girshick, and Jian Sun.
\newblock Faster r-cnn: Towards real-time object detection with region proposal
  networks.
\newblock In {\em NIPS}, 2015.

\bibitem{romero2014fitnets}
Adriana Romero, Nicolas Ballas, Samira~Ebrahimi Kahou, Antoine Chassang, Carlo
  Gatta, and Yoshua Bengio.
\newblock {FitNets}: Hints for thin deep nets.
\newblock In {\em ICLR}, 2014.

\bibitem{russakovsky2015imagenet}
Olga Russakovsky, Jia Deng, Hao Su, Jonathan Krause, Sanjeev Satheesh, Sean Ma,
  Zhiheng Huang, Andrej Karpathy, Aditya Khosla, Michael Bernstein, et~al.
\newblock Imagenet large scale visual recognition challenge.
\newblock {\em IJCV}, 2015.

\bibitem{shen2019meal}
Zhiqiang Shen, Zhankui He, and Xiangyang Xue.
\newblock {MEAL}: Multi-model ensemble via adversarial learning.
\newblock In {\em AAAI}, 2019.

\bibitem{SLJ+15}
Christian Szegedy, Wei Liu, Yangqing Jia, Pierre Sermanet, Scott Reed, Dragomir
  Anguelov, Dumitru Erhan, Vincent Vanhoucke, and Andrew Rabinovich.
\newblock Going deeper with convolutions.
\newblock In {\em CVPR}, 2015.

\bibitem{tarvainen2017mean}
Antti Tarvainen and Harri Valpola.
\newblock Mean teachers are better role models: Weight-averaged consistency
  targets improve semi-supervised deep learning results.
\newblock In {\em NeurIPS}, 2017.

\bibitem{uddin2020saliencymix}
A~F~M Uddin, Mst. Monira, Wheemyung Shin, TaeChoong Chung, and Sung-Ho Bae.
\newblock {SaliencyMix}: A saliency guided data augmentation strategy for
  better regularization.
\newblock In {\em ICML}, 2021.

\bibitem{Vapn99}
VN Vapnik.
\newblock An overview of statistical learning theory.
\newblock {\em Neural Networks, IEEE Transactions on}, 10(5):988--999, 1999.

\bibitem{VSP+17}
Ashish Vaswani, Noam Shazeer, Niki Parmar, Jakob Uszkoreit, Llion Jones,
  Aidan~N Gomez, Lukasz Kaiser, and Illia Polosukhin.
\newblock Attention is all you need.
\newblock In {\em NeurIPS}, 2017.

\bibitem{venkataramanan2021alignmix}
Shashanka Venkataramanan, Ewa Kijak, Laurent Amsaleg, and Yannis Avrithis.
\newblock Alignmixup: Improving representation by interpolating aligned
  features.
\newblock In {\em CVPR}, 2022.

\bibitem{verma2019manifold}
Vikas Verma, Alex Lamb, Christopher Beckham, Amir Najafi, Ioannis Mitliagkas,
  David Lopez-Paz, and Yoshua Bengio.
\newblock Manifold mixup: Better representations by interpolating hidden
  states.
\newblock In {\em ICML}, 2019.

\bibitem{wang2020understanding}
Tongzhou Wang and Phillip Isola.
\newblock Understanding contrastive representation learning through alignment
  and uniformity on the hypersphere.
\newblock In {\em ICML}, 2020.

\bibitem{Wang_2021_CVPR}
Xinlong Wang, Rufeng Zhang, Chunhua Shen, Tao Kong, and Lei Li.
\newblock Dense contrastive learning for self-supervised visual pre-training.
\newblock In {\em CVPR}, 2021.

\bibitem{xiao2010sun}
Jianxiong Xiao, James Hays, Krista~A Ehinger, Aude Oliva, and Antonio Torralba.
\newblock Sun database: Large-scale scene recognition from abbey to zoo.
\newblock In {\em CVPR}, 2010.

\bibitem{xie2020self}
Qizhe Xie, Minh-Thang Luong, Eduard Hovy, and Quoc~V Le.
\newblock Self-training with noisy student improves imagenet classification.
\newblock In {\em CVPR}, 2020.

\bibitem{xie2021propagate}
Zhenda Xie, Yutong Lin, Zheng Zhang, Yue Cao, Stephen Lin, and Han Hu.
\newblock Propagate yourself: Exploring pixel-level consistency for
  unsupervised visual representation learning.
\newblock In {\em CVPR}, 2021.

\bibitem{xie2021simmim}
Zhenda Xie, Zheng Zhang, Yue Cao, Yutong Lin, Jianmin Bao, Zhuliang Yao, Qi
  Dai, and Han Hu.
\newblock Simmim: A simple framework for masked image modeling.
\newblock {\em arXiv preprint arXiv:2111.09886}, 2021.

\bibitem{xiong2021self}
Yuwen Xiong, Mengye Ren, Wenyuan Zeng, and Raquel Urtasun.
\newblock Self-supervised representation learning from flow equivariance.
\newblock In {\em ICCV}, 2021.

\bibitem{yao2015tiny}
Leon Yao and John Miller.
\newblock Tiny imagenet classification with convolutional neural networks.
\newblock Technical report, Standford University, 2015.

\bibitem{yu2015lsun}
Fisher Yu, Ari Seff, Yinda Zhang, Shuran Song, Thomas Funkhouser, and Jianxiong
  Xiao.
\newblock Lsun: Construction of a large-scale image dataset using deep learning
  with humans in the loop.
\newblock {\em arXiv preprint arXiv:1506.03365}, 2015.

\bibitem{yun2019cutmix}
Sangdoo Yun, Dongyoon Han, Seong~Joon Oh, Sanghyuk Chun, Junsuk Choe, and
  Youngjoon Yoo.
\newblock Cutmix: Regularization strategy to train strong classifiers with
  localizable features.
\newblock In {\em ICCV}, 2019.

\bibitem{zagoruyko2016paying}
Sergey Zagoruyko and Nikos Komodakis.
\newblock Paying more attention to attention: Improving the performance of
  convolutional neural networks via attention transfer.
\newblock In {\em ICLR}, 2016.

\bibitem{zagoruyko2016wide}
Sergey Zagoruyko and Nikos Komodakis.
\newblock Wide residual networks.
\newblock In {\em BMVC}, 2016.

\bibitem{zhang2018mixup}
Hongyi Zhang, Moustapha Cisse, Yann~N Dauphin, and David Lopez-Paz.
\newblock mixup: Beyond empirical risk minimization.
\newblock In {\em ICLR}, 2018.

\bibitem{zhang2019your}
Linfeng Zhang, Jiebo Song, Anni Gao, Jingwei Chen, Chenglong Bao, and Kaisheng
  Ma.
\newblock Be your own teacher: Improve the performance of convolutional neural
  networks via self distillation.
\newblock In {\em ICCV}, 2019.

\bibitem{zhang2018adversarial}
Xiaolin Zhang, Yunchao Wei, Jiashi Feng, Yi Yang, and Thomas~S Huang.
\newblock Adversarial complementary learning for weakly supervised object
  localization.
\newblock In {\em CVPR}, 2018.

\bibitem{zhou2016learning}
Bolei Zhou, Aditya Khosla, Agata Lapedriza, Aude Oliva, and Antonio Torralba.
\newblock Learning deep features for discriminative localization.
\newblock In {\em CVPR}, 2016.

\bibitem{zhou2021ibot}
Jinghao Zhou, Chen Wei, Huiyu Wang, Wei Shen, Cihang Xie, Alan Yuille, and Tao
  Kong.
\newblock {iBOT}: Image bert pre-training with online tokenizer.
\newblock In {\em ICLR}, 2022.

\bibitem{zhu2020automix}
Jianchao Zhu, Liangliang Shi, Junchi Yan, and Hongyuan Zha.
\newblock Automix: Mixup networks for sample interpolation via cooperative
  barycenter learning.
\newblock In {\em ECCV}, 2020.

\end{thebibliography}
\bibliographystyle{ieee_fullname}

\clearpage

\setcounter{page}{1}

\appendix


\section{More experiments}
\label{sec:more-exp}

\subsection{More on setup}
\label{sec:more-setup}

\paragraph{Settings and hyperparameters}

We train MultiMix and its variants with mixed examples only. We use a mini-batch of size $m = 128$ examples in all experiments. For every mini-batch, we apply MultiMix with probability $0.5$ or input mixup otherwise. For input mixup, we interpolate the standard $m$ pairs~\eq{input}. For MultiMix, we use the entire network as the encoder $f_\theta$ by default, except for the last fully-connected layer, which we use as classifier $g_W$. We use $n = 1000$ tuples and draw a different $\alpha \sim U[0.5, 2.0]$ for each example from the Dirichlet distribution by default. For multi-GPU experiments, all training hyperparameters including $m$ and $n$ are per GPU.

For dense MultiMix, the spatial resolution is $4 \times 4$ ($r = 16$) on CIFAR-10/100 and $7 \times 7$ ($r = 49$) on Imagenet by default. We obtain the attention map by~\eq{attn} using GAP for vector $u$ and ReLU followed by $\ell_1$ normalization as non-linearity $h$ by default. To predict class probabilities and compute the loss densely, we use the classifier $g_W$ as $1 \times 1$ convolution by default; when interpolating at earlier layers, we follow the process described in \autoref{sec:dense}. For distillation, both the teacher and student networks have the same architecture. By default, we use $\gamma = \frac{1}{2}$ in~\eq{distill-1}, that is, equal contribution of original labels and teacher predictions.

\paragraph{CIFAR-10/100 training}

Following the experimental settings of AlignMixup~\citep{venkataramanan2021alignmix}, we train MultiMix and its variants using SGD for $2000$ epochs using the same random seed as AlignMixup. We set the initial learning rate to $0.1$ and decay it by a factor of $0.1$ every $500$ epochs. The momentum is set to $0.9$ and the weight decay to $0.0001$. We use a batch size $m = 128$ and train on a single NVIDIA RTX 2080 TI GPU for $10$ hours.

\paragraph{TinyImageNet training}

Following the experimental settings of PuzzleMix~\citep{kim2020puzzle}, we train MultiMix and its variants using SGD for $1200$ epochs, using the same random seed as AlignMixup. We set the initial learning rate to $0.1$ and decay it by a factor of $0.1$ after $600$ and $900$ epochs. The momentum is set to $0.9$ and the weight decay to $0.0001$. We train on two NVIDIA RTX 2080 TI GPUs for $18$ hours.

\paragraph{ImageNet training}

Following the experimental settings of PuzzleMix~\citep{kim2020puzzle}, we train MultiMix and its variants using the same random seed as AlignMixup. We train R-50 using SGD with momentum 0.9 and weight decay $0.0001$ and ViT-S/16 using AdamW with default parameters. The initial learning rate is set to $0.1$ and $0.01$, respectively. We decay the learning rate by $0.1$ at $100$ and $200$ epochs. We train on 32 NVIDIA V100 GPUs for $20$ hours.

\paragraph{Tasks and metrics}

We use top-1 error (\%, lower is better) or top-1 accuracy (\%, higher is better) as evaluation metric on \emph{image classification} and \emph{robustness to adversarial attacks} (\autoref{sec:class} and \autoref{sec:more-class}). Additional datasets and metrics are reported separately for \emph{transfer learning to object detection} (\autoref{sec:det}) and \emph{out-of-distribution detection} (\autoref{sec:more-ood}).


\subsection{More results: Classification and robustness}
\label{sec:more-class}

Using the experimental settings of~\autoref{sec:more-setup}, we extend~\autoref{tab:cls} and~\autoref{tab:adv} of~\autoref{sec:class} by comparing MultiMix and its variants with additional mixup methods in~\autoref{tab:more-cls} and~\autoref{tab:more-adv}. The additional methods are Input mixup~\citep{zhang2018mixup}, Cutmix~\citep{yun2019cutmix}, SaliencyMix~\citep{uddin2020saliencymix}, StyleMix~\citep{hong2021stylemix}, StyleCutMix~\citep{hong2021stylemix} and SuperMix~\citep{Dabouei_2021_CVPR}. We reproduce SuperMix using the official code\footnote{\url{https://github.com/alldbi/SuperMix}}, which first trains the teacher network  using clean examples and then the student using mixed. For fair comparison, we use the same network as the teacher and student models.

In \autoref{tab:more-cls} and \autoref{tab:more-adv}, we observe that MultiMix and its variants outperform all the additional mixup methods on image classification. Furthermore, they are more robust to FGSM and PGD attacks as compared to these additional methods. The remaining observations in \autoref{sec:class} are still valid.

\begin{table}
\begin{subtable}[T]{0.51\textwidth}
\centering
\scriptsize
\setlength{\tabcolsep}{4.5pt}
\begin{tabular}{lccccc} \toprule
	\Th{Dataset}                                            & \mc{2}{\Th{Cifar-10}}             & \mc{2}{\Th{Cifar-100}}            & TI              \\
	\Th{Network}                                            & R-18            & W16-8           & R-18            & W16-8           & R-18            \\ \midrule
	Baseline$^\dagger$                                      & 5.19            & 5.11            & 23.24           & 20.63           & 43.40           \\
	Input mixup~\citep{zhang2018mixup}$^\dagger$            & 4.03            & 3.98            & 20.21           & 19.88           & 43.48           \\
	CutMix~\citep{yun2019cutmix}$^\dagger$                  & 3.27            & 3.54            & 19.37           & 19.71           & 43.11           \\
	Manifold mixup~\citep{verma2019manifold}$^\dagger$      & 2.95            & 3.56            & 19.80           & 19.23           & 40.76           \\
	PuzzleMix~\citep{kim2020puzzle}$^\dagger$               & 2.93            & \ul{2.99}       & 20.01           & 19.25           & 36.52           \\
	Co-Mixup~\citep{kim2021co}$^\dagger$                    & \ul{2.89}       & 3.04            & 19.81           & 19.57           & 35.85           \\
	SaliencyMix~\citep{uddin2020saliencymix}$^\dagger$      & 2.99            & 3.53            & 19.69           & 19.59           & 33.81           \\
	StyleMix~\citep{hong2021stylemix}$^\dagger$             & 3.76            & 3.89            & 20.04           & 20.45           & 36.13           \\
	StyleCutMix~\citep{hong2021stylemix}$^\dagger$          & 3.06            & 3.12            & 19.34           & 19.28           & 33.49           \\
	SuperMix~\citep{Dabouei_2021_CVPR}$^*$                  & 4.01            & 3.87            & 20.99           & 20.13           & 38.55           \\
	AlignMixup~\citep{venkataramanan2021alignmix}$^\dagger$ & 2.95            & 3.09            & \ul{18.29}      & \ul{18.77}      & \ul{33.13}      \\ \midrule
	MultiMix (ours)                                         & 2.97            & 2.92            & 18.19           & 18.57           & 32.78           \\
	\hspace{3pt}$+$ distil                                  & \se{2.87}       & \tb{2.78}       & \se{17.72}      & \se{17.91}      & 31.93           \\
	\hspace{3pt}$+$ dense                                   & 2.91            & \se{2.89}       & 18.12           & 18.20           & \se{31.54}           \\
	\rowcolor{LightCyan}
	\hspace{3pt}$+$ dense $+$ distil                        & \tb{2.81}       & \tb{2.78}       & \tb{17.48}      & \tb{17.66}      & \tb{30.87}      \\ \midrule
	Gain                                                    & \gp{\tb{+0.08}} & \gp{\tb{+0.21}} & \gp{\tb{+0.81}} & \gp{\tb{+1.11}} & \gp{\tb{+2.26}} \\ \bottomrule
\end{tabular}
\caption{\emph{Image classification} top-1 error (\%) on CIFAR-10/100 and TI (TinyImagenet). R: PreActResnet, W: WRN.}
\end{subtable}
\hspace{6pt}
\begin{subtable}[T]{0.45\textwidth}
\centering
\scriptsize
\setlength{\tabcolsep}{2pt}
\begin{tabular}{lcccc} \toprule
	\Th{Network}                                              & \mc{2}{\Th{Resnet-50}}            & \mc{2}{\Th{ViT-S/16}}              \\
	\Th{Method}                                               & \Th{Speed}      & \Th{Error}      & \Th{Speed}      & \Th{Error}       \\ \midrule
	Baseline$^\dagger$                                        & 1.17            & 23.68           & 1.01            & 26.1             \\
	Input mixup~\citep{zhang2018mixup}$^\dagger$              & 1.14            & 22.58           & 0.99            & 25.3             \\
	CutMix~\citep{yun2019cutmix}$^\dagger$                    & 1.16            & 21.40           & 0.99            & 25.6             \\
	Manifold mixup~\citep{verma2019manifold}$^\dagger$        & 1.15            & 22.50           & 0.97            & 24.8             \\
	PuzzleMix~\citep{kim2020puzzle}$^\dagger$                 & 0.84            & 21.24           & 0.73            & 24.3             \\
	Co-Mixup~\citep{kim2021co}$^\dagger$                      & 0.62            & --              & 0.57            & \ul{24.1}        \\
	SaliencyMix~\citep{uddin2020saliencymix}$^\dagger$        & 1.14            & 21.26           & 0.96            & 24.2             \\
	StyleMix~\citep{hong2021stylemix}$^\dagger$               & 0.99            & 24.06           & 0.85            & 25.2             \\
	StyleCutMix~\citep{hong2021stylemix}$^\dagger$            & 0.76            & 22.71           & 0.71            & 24.9             \\
	SuperMix~\citep{Dabouei_2021_CVPR}$^*$                    & 0.92            & 22.40           & --              & --               \\
	AlignMixup~\citep{venkataramanan2021alignmix}$^\dagger$   & 1.03            & \ul{20.68}      & --              & --               \\ \midrule
	MultiMix (ours)                                           & 1.16            & 21.19           & 1.0             & 24.8             \\
	\hspace{3pt}$+$ distil                                    & 1.06            & \se{19.88}      & 0.93            & \se{23.4}        \\
	\hspace{3pt}$+$ dense                                     & 0.95            & 20.63           & 0.88            & 23.9             \\
	\rowcolor{LightCyan}
	\hspace{3pt}$+$ dense $+$ distil                          & 0.83            & \tb{19.79}      & 0.81            & \tb{23.1}        \\ \midrule
	Gain                                                      &                 & \gp{\tb{+0.89}} &                 & \gp{\tb{+1.0}}   \\ \bottomrule
\end{tabular}
\caption{\emph{Image classification and training speed} on ImageNet. Top-1 error (\%): lower is better. Speed: images/sec ($\times 10^3$): higher is better.}
\end{subtable}
\caption{\emph{Image classification and training speed.} $^\dagger$: reported by AlignMixup. $^*$: reproduced, same teacher and student model. \tb{Bold black}: best; \se{Blue}: second best; underline: best baseline. Gain: reduction of error over best baseline.}
\label{tab:more-cls}
\end{table}

\begin{table}
\centering
\scriptsize
\setlength{\tabcolsep}{6pt}
\begin{tabular}{lccccc|cccc} \toprule
	\Th{Attack}                                             & \multicolumn{5}{c}{FGSM}                                                                & \multicolumn{4}{c}{PGD}                                               \\ \midrule
	\Th{Dataset}                                            & \mc{2}{\Th{Cifar-10}}             & \mc{2}{\Th{Cifar-100}}            & TI              & \mc{2}{\Th{Cifar-10}}             & \mc{2}{\Th{Cifar-100}}            \\
	\Th{Network}                                            & R-18            & W16-8           & R-18            & W16-8           & R-18            & R-18            & W16-8           & R-18            & W16-8           \\ \midrule
	Baseline$^\dagger$                                      & 89.41           & 88.02           & 87.12           & 72.81           & 91.85           & 99.99           & 99.94           & 99.97           & 99.99           \\
	Input mixup~\citep{zhang2018mixup}$^\dagger$            & 78.42           & 79.21           & 81.30           & 67.33           & 88.68           & 99.77           & 99.43           & 99.96           & 99.37           \\
	CutMix~\citep{yun2019cutmix}$^\dagger$                  & 77.72           & 78.33           & 86.96           & 60.16           & 88.68           & 99.82           & 98.10           & 98.67           & 97.98           \\
	Manifold mixup~\citep{verma2019manifold}$^\dagger$      & 77.63           & 76.11           & 80.29           & 56.45           & 89.25           & 97.22           & 98.49           & 99.66           & 98.43           \\
	PuzzleMix~\citep{kim2020puzzle}$^\dagger$               & 57.11           & 60.73           & 78.70           & 57.77           & 83.91           & 97.73           & 97.00           & 96.42           & 95.28           \\
	Co-Mixup~\citep{kim2021co}$^\dagger$                    & 60.19           & 58.93           & 77.61           & 56.59           & --              & 97.59           & \ul{96.19}      & 95.35           & 94.23           \\
	SaliencyMix~\citep{uddin2020saliencymix}$^\dagger$      & 57.43           & 68.10           & 77.79           & 58.10           & 81.16           & 97.51           & 97.04           & 95.68           & 93.76           \\
	StyleMix~\citep{hong2021stylemix}$^\dagger$             & 79.54           & 71.05           & 80.54           & 67.94           & 84.93           & 98.23           & 97.46           & 98.39           & 98.24           \\
	StyleCutMix~\citep{hong2021stylemix}$^\dagger$          & 58.79           & \ul{56.12}      & 77.49           & 56.83           & 80.59           & 97.87           & 96.70           & 91.88           & 93.78           \\
	SuperMix~\citep{Dabouei_2021_CVPR}$^*$                  & 59.98           & 58.10           & 78.75           & 58.19           & 81.03           & 97.65           & 97.20           & 91.51           & 92.73           \\
	AlignMixup~\citep{venkataramanan2021alignmix}$^\dagger$ & \ul{54.83}      & 56.20           & \ul{74.18}      & \ul{55.05}      & \ul{78.83}      & \ul{95.42}      & 96.71           & \ul{90.40}      & \ul{92.16}      \\ \midrule
	MultiMix (ours)                                         & 54.19           & 55.39           & 75.84           & 54.58           & 77.51           & 94.27           & 94.83           & 90.02           & 91.68           \\
	\hspace{3pt}$+$ distillation                            & \se{52.55}      & \se{51.42}      & \se{73.55}      & \se{52.77}      & 76.20           & \se{92.69}      & 93.90           & 88.87           & \se{90.54}      \\
	\hspace{3pt}$+$ dense                                   & 54.10           & 53.33           & 74.48           & 53.01           & \se{75.57}      & 92.99           & \se{92.68}      & \se{88.60}      & 90.90           \\
	\rowcolor{LightCyan}
	\hspace{3pt}$+$ dense $+$ distillation                  & \tb{52.07}      & \tb{50.17}      & \tb{72.98}      & \tb{52.19}      & \tb{75.18}      & \tb{90.82}      & \tb{90.56}      & \tb{87.58}      & \tb{90.18}      \\ \midrule
	Gain                                                    & \gp{\tb{+2.76}} & \gp{\tb{+5.95}} & \gp{\tb{+1.20}} & \gp{\tb{+2.86}} & \gp{\tb{+3.65}} & \gp{\tb{+4.60}} & \gp{\tb{+5.63}} & \gp{\tb{+2.82}} & \gp{\tb{+1.98}}  \\ \bottomrule
\end{tabular}
\vspace{6pt}
\caption{\emph{Robustness to FGSM \& PGD attacks}. Top-1 error (\%): lower is better. $^\dagger$: reported by AlignMixup. $^*$: reproduced, same teacher and student model. \tb{Bold black}: best; \se{Blue}: second best; underline: best baseline. Gain: reduction of error over best baseline. TI: TinyImagenet. R: PreActResnet, W: WRN.}
\label{tab:more-adv}
\end{table}

\subsection{More results: Out of distribution detection}
\label{sec:more-ood}

\begin{table}
\centering
\scriptsize
\setlength{\tabcolsep}{3pt}
\begin{tabular}{lcccc|cccc|cccc}
\toprule
	\Th{Task}                                                & \mc{11}{\Th{Out-Of-Distribution Detection}}                                                                                                                                                              \\ \midrule
	\Th{Dataset}                                             & \mc{4}{\Th{LSUN (crop)}}                                              & \mc{4}{\Th{iSUN}}                                                      & \mc{4}{\Th{TI (crop)}}                                           \\ \midrule
	\mr{2}{\Th{Metric}}                                      & \Th{Det}        & \Th{AuROC}      & \Th{AuPR}      & \Th{AuPR}        & \Th{Det}        & \Th{AuROC}      & \Th{AuPR}       & \Th{AuPR}        & \Th{Det}       & \Th{AuROC}     & \Th{AuPR}      & \Th{AuPR}     \\
																				& \Th{Acc}        &                 & (ID)           & (OOD)            & \Th{Acc}        &                 & (ID)            & (OOD)            & \Th{Acc}       &                & (ID)           & (OOD)         \\ \midrule
	Baseline$^\dagger$                                       & 54.0            & 47.1            & 54.5           & 45.6             & 66.5            & 72.3            & 74.5            & 69.2             & 61.2           & 64.8            & 67.8           & 60.6         \\
	Input mixup~\citep{zhang2018mixup}$^\dagger$             & 57.5            & 59.3            & 61.4           & 55.2             & 59.6            & 63.0            & 60.2            & 63.4             & 58.7           & 62.8            & 63.0           & 62.1         \\
	Cutmix~\citep{yun2019cutmix}$^\dagger$                   & 63.8            & 63.1            & 61.9           & 63.4             & 67.0            & 76.3            & 81.0            & 77.7             & 70.4           & 84.3            & 87.1           & 80.6         \\
	Manifold mixup~\citep{verma2019manifold}$^\dagger$       & 58.9            & 60.3            & 57.8           & 59.5             & 64.7            & 73.1            & 80.7            & 76.0             & 67.4           & 69.9            & 69.3           & 70.5         \\
	PuzzleMix~\citep{kim2020puzzle}$^\dagger$                & 64.3            & 69.1            & 80.6           & 73.7             & \ul{73.9}       & 77.2            & 79.3            & 71.1             & 71.8           & 76.2            & 78.2           & 81.9         \\
	Co-Mixup~\citep{kim2021co}$^\dagger$                     & 70.4            & 75.6            & 82.3           & 70.3             & 68.6            & 80.1            & 82.5            & 75.4             & 71.5           & 84.8            & 86.1           & 80.5         \\
	SaliencyMix~\citep{uddin2020saliencymix}$^\dagger$       & 68.5            & 79.7            & 82.2           & 64.4             & 65.6            & 76.9            & 78.3            & 79.8             & 73.3           & 83.7            & 87.0           & 82.0         \\
	StyleMix~\citep{hong2021stylemix}$^\dagger$              & 62.3            & 64.2            & 70.9           & 63.9             & 61.6            & 68.4            & 67.6            & 60.3             & 67.8           & 73.9            & 71.5           & 78.4         \\
	StyleCutMix~\citep{hong2021stylemix}$^\dagger$           & 70.8            & 78.6            & 83.7           & 74.9             & 70.6            & 82.4            & 83.7            & 76.5             & 75.3           & 82.6            & 82.9           & 78.4         \\
	SuperMix~\citep{Dabouei_2021_CVPR}$^*$                   & 70.9            & 77.4            & 80.1           & 72.3             & 71.0            & 76.8            & 79.6            & 76.7             & 75.1           & 82.8            & 82.5           & 78.6         \\
	AlignMixup~\citep{venkataramanan2021alignmix}$^\dagger$  & \ul{74.2}       & \ul{79.9}       & \ul{84.1}      & \ul{75.1}        & 72.8            & \ul{83.2}       & \ul{84.1}       & \ul{80.3}        & \ul{77.2}     & \ul{85.0}       & \ul{87.8}      & \ul{85.0}         \\ \midrule
	MultiMix (ours)                                          & 79.2            & 82.6            & 85.2           & 77.6             & 75.6            & 85.1            & 87.8            & 83.1             & 78.3           & 86.6             & 89.0            & \se{88.2}      \\
	\hspace{3pt}$+$ distillation                             & 80.3            & \se{84.4}       & \se{86.3}      & 76.4             & \se{79.0}       & \se{85.6}       & \se{88.2}       & \tb{84.9}        & 80.7           & 87.8             & 89.9            & \se{88.2}       \\
	\hspace{3pt}$+$ dense                                    & \se{80.8}       & 84.3            & 85.9           & \se{78.0}        & 76.8            & 85.4            & 88.0            & 84.6             & \se{81.4}      & \se{89.0}        & \tb{90.8}       & 88.0        \\
	\rowcolor{LightCyan}
	\hspace{3pt}$+$ dense $+$ distillation                   & \tb{81.0}       & \tb{84.9}       & \tb{86.4}       & \tb{78.2}       & \tb{79.2}       & \tb{86.0}       & \tb{88.5}       & \se{84.8}        & \tb{81.9}      & \tb{89.3}        & \se{90.3}      & \tb{88.3}   \\ \midrule
	Gain                                                     & \gp{\tb{+6.8}}  & \gp{\tb{+5.0}}  & \gp{\tb{+2.3}}  & \gp{\tb{+3.1}}  & \gp{\tb{+5.3}}  & \gp{\tb{+2.8}}  & \gp{\tb{+4.4}}  & \gp{\tb{+4.6}}   & \gp{\tb{+4.7}} & \gp{\tb{+4.3}}   & \gp{\tb{+3.0}} & \gp{\tb{+3.3}} \\ \bottomrule
\end{tabular}
\vspace{6pt}
\caption{\emph{Out-of-distribution detection} using R-18. Det Acc (detection accuracy), AuROC, AuPR (ID) and AuPR (OOD): higher is better. $^\dagger$: reported by AlignMixup. $^*$: reproduced, same teacher and student model. \tb{Bold black}: best; \se{Blue}: second best; underline: best baseline. Gain: increase in performance. TI: TinyImagenet.}
\label{tab:more-ood}
\end{table}

This is a standard benchmark for evaluating over-confidence. Here, \emph{in-distribution} (ID) are examples on which the network has been trained, and \emph{out-of-distribution} (OOD) are examples drawn from any other distribution. Given a mixture of ID and OOD examples, the network should predict an ID example with high confidence and an OOD example with low confidence, \ie, the confidence of the predicted class should be below a certain threshold.

Following AlignMixup~\citep{venkataramanan2021alignmix}, we compare MultiMix and its variants with SoTA methods trained using R-18 on CIFAR-100 as ID examples, while using LSUN~\citep{yu2015lsun}, iSUN~\citep{xiao2010sun} and TI to draw OOD examples. We use detection accuracy, Area under ROC curve (AuROC) and Area under precision-recall curve (AuPR) as evaluation metrics. In \autoref{tab:more-ood}, we observe that MultiMix and its variants outperform SoTA on all datasets and metrics by a large margin. Although the gain of vanilla MultiMix and Dense MultiMix over SoTA mixup methods is small on image classification, these variants significantly reduce over-confident incorrect predictions and achieve superior performance on out-of-distribution detection.


\subsection{More ablations}
\label{sec:more-ablation}

As in \autoref{sec:ablation}, all ablations here are performed using R-18 on CIFAR-100.

\paragraph{Mixup methods with distillation}

In \autoref{sec:class} and \autoref{tab:more-cls}, we observe that distillation significantly improves the performance when used with MultiMix. Here, we also study its effect when applied to SoTA mixup methods.

Given a mini-batch of $m$ examples with inputs $X$ and targets $Y$, we obtain the augmented views $V$ and $V'$ as discussed in \autoref{sec:distill}. We then follow the mixup strategy of each mixup method and obtain the corresponding predicted class probabilities $\wt{P}, \wt{P}'$ from the student and teacher classifier, respectively. \Eg, for manifold mixup~\citep{verma2019manifold}, we interpolate the embeddings $Z = f_\theta(V), Z' = f_{\theta'}(V')$ using \eq{embed} and obtain $\wt{P} = g_W(\wt{Z})$ and $\wt{P}' = g_{W'}(\wt{Z}')$. In each case, we obtain the interpolated targets $\wt{Y}$ using \eq{target} and train the student network using \eq{distill-1}.

In \autoref{tab:baseline-distil-dense}, we observe that with distillation, the performance of all SoTA mixup methods improve. For example, the baseline improves by 1.52\% accuracy (76.76 $\rightarrow$ 78.23) and manifold mixup by 1.12\% (80.20 $\rightarrow$ 81.32). On average, we observe a gain of 1\% brought by distillation. An exception is AlignMixup~\citep{venkataramanan2021alignmix}: distillation brings a marginal improvement of 0.09\% (81.71 $\rightarrow$ 81.80), making it on-par with vanilla MultiMix.


\paragraph{Mixup methods with dense loss}

In \autoref{tab:more-cls} we observe that dense interpolation and dense loss improve vanilla MultiMix. Here, we study the effect of the dense loss when applied to SoTA mixup methods.

Given a mini-batch of $m$ examples, we follow the mixup strategy of the SoTA mixup methods to obtain the mixed embedding $\wt{Z}^j \in \real^{d \times m}$ for each spatial position $j = 1, \dots, r$. Then, as discussed in \autoref{sec:dense}, we obtain the predicted class probabilities $\wt{P}^j \in \real^{c \times m}$ again for each $j = 1, \dots, r$. Finally, we compute the cross-entropy loss $H(\wt{Y}, \wt{P}^j)$~\eq{ce} densely at each spatial position $j$, where the interpolated target label $\wt{Y} \in \real^{c \times m}$ is given by~\eq{target}.

In \autoref{tab:baseline-distil-dense}, we observe that using a dense loss improves the performance of all SoTA mixup methods. The baseline improves by 1.4\% accuracy (76.76 $\rightarrow$ 78.16) and manifold mixup by 0.67\% (80.20 $\rightarrow$ 80.87). On average, we observe a gain of 0.7\% brought by the dense loss. An exception is AlignMixup~\citep{venkataramanan2021alignmix}, which drops by 0.35\% (81.71 $\rightarrow$ 81.36). This may be due to the alignment process, whereby the interpolated dense embeddings are not very far from the original.

Finally, we study the effect of using a dense distillation loss on SoTA mixup methods. Here, similarly with~\eq{distill-1}, the loss has two terms for each spatial position $j$: the first is the dense cross-entropy loss $H(\wt{Y}, \wt{P}^j)$ as above and the second is the dense distillation loss $H((\wt{P}')^j, \wt{P}^j)$, where $\wt{P}'$ is obtained by the teacher. In \autoref{tab:baseline-distil-dense}, we observe that dense distillation further improves the performance of SoTA mixup methods as compared to using the dense loss only.

\begin{table}
\centering
\scriptsize
\begin{tabular}{lcccc} \toprule
	\Th{Method}                                   & \Th{Vanilla}  & \Th{$+$ Distil}  & \Th{$+$ Dense}  & \Th{$+$ Dense $+$ Distil}  \\ \midrule
	Baseline                                      & 76.76         & 78.28            & 78.16           & 79.07                      \\
	Input mixup~\citep{zhang2018mixup}            & 79.79         & 80.19            & 80.21           & 80.54                      \\
	CutMix~\citep{yun2019cutmix}                  & 80.63         & 81.51            & 81.40           & 81.61                      \\
	Manifold mixup~\citep{verma2019manifold}      & 80.20         & 81.32            & 80.87           & 81.47                      \\
	PuzzleMix~\citep{kim2020puzzle}               & 79.99         & 81.26            & 80.62           & 81.44                      \\
	Co-Mixup~\citep{kim2021co}                    & 80.19         & 81.39            & 80.84           & 81.69                      \\
	SaliencyMix~\citep{uddin2020saliencymix}      & 80.31         & 81.57            & 81.21           & 81.73                      \\
	StyleMix~\citep{hong2021stylemix}             & 79.96         & 81.22            & 80.76           & 81.30                      \\
	StyleCutMix~\citep{hong2021stylemix}          & 80.66         & 81.60            & 81.41           & 81.75                      \\
	SuperMix~\citep{Dabouei_2021_CVPR}$^*$        & 79.01         & 80.83            & 80.12           & 80.83                      \\
	AlignMixup~\citep{venkataramanan2021alignmix} & 81.71         & 81.80            & 81.36           & 81.40                      \\ \midrule
	MultiMix (ours)$^\ddagger$                    & --            & --               & 81.84           & 82.30                      \\
	\rowcolor{LightCyan}
	MultiMix (ours)                               & \tb{81.81}    & \tb{82.28}       & \tb{81.88}      & \tb{82.52}                 \\ \bottomrule
\end{tabular}
\vspace{6pt}
\caption{\emph{Image classification} on CIFAR-100 using R-18: The effect of distillation, dense loss and both on SoTA mixup methods. Top-1 accuracy (\%): higher is better. $^*$: `vanilla' refers to teacher pre-training and `distil' to self-distillation where teacher and student are trained concurrently from scratch. $^\ddagger$: Instead of dense MultiMix, we only apply the loss densely.}
\label{tab:baseline-distil-dense}
\end{table}

\begin{table}
\centering
\scriptsize
\begin{tabular}{lcccccc} \toprule
	\Th{Method}                 & $u$ & $h$                  & --         & \Th{$+$Distil}  \\ \midrule
	Uniform                     & --  & --                   & 81.33      & 81.59           \\ \midrule
	\mr{4}{Attention~\eq{attn}} & CAM & softmax              & 81.21      & 81.45           \\
	                            & CAM & $\ell_1 \circ \relu$ & 81.63      & 81.91           \\ \cmidrule{2-5}
	                            & GAP & softmax              & 81.78      & 82.01           \\
	\rowcolor{LightCyan}
	                            & GAP & $\ell_1 \circ \relu$ & \tb{81.88} & \tb{82.52}      \\ \bottomrule
\end{tabular}
\vspace{6pt}
\caption{\emph{Variants of spatial attention} in dense MultiMix, with and without distillation, on CIFAR-100 using R-18. Top-1 accuracy (\%): higher is better. GAP: Global Average Pooling; CAM: Class Activation Maps~\citep{zhou2016learning}; $\ell_1 \circ \relu$: ReLU followed by $\ell_1$ normalization.}
\label{tab:dense-attn}
\end{table}

\paragraph{Two-stage distillation}

Following SuperMix~\cite{Dabouei_2021_CVPR}, we also study the effect of using a two-stage distillation process with MultiMix, rather than online self-distillation.

In the first stage, we train the teacher using only clean examples for 300 epochs, and we achieve a top-1 accuracy of 75.62\%. This is slightly lower than the 76.76\% of the baseline from \autoref{tab:cls}(a), which is trained for 2000 epochs. In the second stage, we fix the teacher parameters and train the student using the predictions from the teacher network as targets. In particular, we use the second term $H(\wt{P}', \wt{P})$ of~\eq{distill-1}, that is, $\gamma = 0$. At inference, the top-1 accuracy drops by 16\% (75.62 $\rightarrow$ 59.77). This shows that using the setting of SuperMix is not effective, while also being computationally expensive because of the two-stage training.

We also study the effect of training the student with both the interpolated labels $\wt{Y}$\eq{multi-target} and the interpolated predictions $\wt{P}'$ of the pretrained teacher as targets. In particular, we use~\eq{distill-1} with our default $\gamma = \frac{1}{2}$. At inference, the top-1 accuracy improves by 4.7\% compared with the teacher (75.62 $\rightarrow$ 80.35). However, the student accuracy of 80.35\% is still inferior to our 82.28\% by online self-distillation (\autoref{tab:baseline-distil-dense}). This shows that joint training of teacher and student is beneficial.

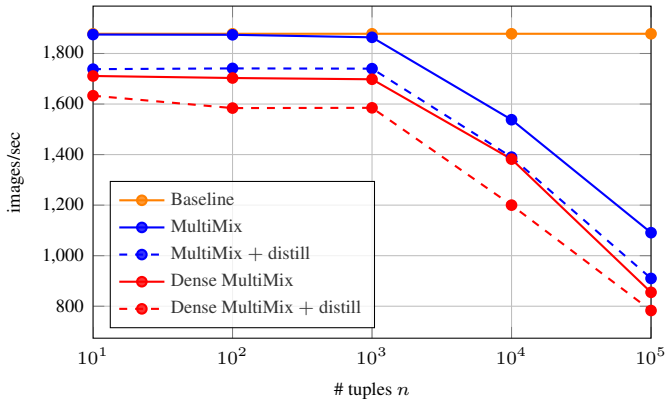
\begin{figure}[!h]
\centering
\begin{tikzpicture}
\begin{axis}[
	width=9cm,
	height=6cm,
	font=\scriptsize,
	ylabel={images/sec},
	xlabel={\# tuples $n$},
	enlarge x limits=false,
	xtick={0,...,4},
	xticklabels={$10^1$,$10^2$,$10^3$, $10^4$, $10^5$},
	legend pos=south west,
]
	\pgfplotstableread{
		x    base    mm     mm-d   dmm    dmm-d
		0    1878    1875   1738   1711   1633
		1    1878    1874   1741   1703   1584
		2    1878    1864   1740   1698   1585
		3    1878    1538   1390   1382   1200
		4    1878    1091    910    855    783
	}{\sp}
	\addplot[orange,mark=*] table[y=base]        {\sp}; \leg{Baseline}
	\addplot[blue,mark=*] table[y=mm]            {\sp}; \leg{MultiMix}
	\addplot[blue,mark=*,dashed] table[y=mm-d]   {\sp}; \leg{MultiMix $+$ distill}
	\addplot[red,mark=*] table[y=dmm]            {\sp}; \leg{Dense MultiMix}
	\addplot[red,mark=*,dashed] table[y=dmm-d]   {\sp}; \leg{Dense MultiMix $+$ distill}
\end{axis}
\end{tikzpicture}
\caption{\emph{Training speed} (images/sec) of MultiMix and its variants \vs number of tuples $n$ on CIFAR-100 using R-18. Measured on NVIDIA RTX 2080 TI GPU, including forward and backward pass.}
\vspace{-6pt}
\label{fig:speed-plot}
\end{figure}

\paragraph{Training speed}

In \autoref{fig:speed-plot}, we analyze the training speed of MultiMix and its variants as a function of number of tuples $n$. In terms of speed, vanilla MultiMix is on par with the baseline up to $n=1000$, while bringing an accuracy gain of $5\%$. The best performing variant---dense MultiMix with distillation---is only slower by $15.6\%$ at $n=1000$ as compared to the baseline, which is arguably worth given the impressive $5.8\%$ accuracy gain. Further increasing beyond $n > 1000$ brings a drop in training speed, due to computing $\Lambda$ and then using it to interpolate~\eq{multi-embed},\eq{multi-target}. Because $n > 1000$ also brings little performance benefit according to \autoref{fig:ablation-plot}(b), we set $n = 1000$ as default for all MultiMix variants.


\paragraph{Dense MultiMix: Spatial attention}

In \autoref{sec:dense}, we discuss different options for attention in dense MultiMix. In particular, no attention amounts to defining a uniform $a = \vone_r / r$. Otherwise, $a$ is defined by~\eq{attn}. The vector $u$ can be defined as $u = \vz \vone_r / r$ by global average pooling (GAP) of $\vz$, which is the default, or $u = W y$ assuming a linear classifier with $W \in \real^{d \times c}$. The latter is similar to class activation mapping (CAM)~\cite{zhou2016learning}, but here the current value of $W$ is used online while training. The non-linearity $h$ can be softmax or ReLU followed by $\ell_1$ normalization ($\ell_1 \circ \relu$), which is the default. Here, we study the affect of these options on the performance of dense Multimix.

In \autoref{tab:dense-attn}, we observe that using GAP for $u$ and $\ell_1 \circ \relu$ as $h$ yields the best performance overall. Changing GAP to CAM or $\ell_1 \circ \relu$ to softmax is inferior, more so in the presence of distillation. The combination of CAM with softmax is the weakest, even weaker than uniform attention. CAM may fail because of using the non-optimal value of $W$ while training; softmax may fail because of being too selective. Compared to our best result, uniform attention is clearly inferior, by nearly 1\% in the presence of distillation. This validates that the use of spatial attention in dense MultiMix is clearly beneficial. The intuition is the same as in weakly supervised tasks: In the absence of dense targets, assuming the same target of the entire example at every spatial position naively implies that the object of interest is present everywhere, whereas spatial attention provides a better hint as to where the object may really be.


\paragraph{Dense MultiMix: Spatial resolution}

We study the effect of spatial resolution on dense MultiMix. By default, we use a resolution of $4 \times 4$ at the last residual block of R-18 on CIFAR-100. Here, we additionally investigate $1 \times 1$ (downsampling by average pooling with kernel size 4, same as GAP), $2 \times 2$ (downsampling by average pooling with kernel size $2$) and $8 \times 8$ (upsampling by using stride $1$ in the last residual block). We measure accuracy 81.07\% for spatial resolution $1 \times 1$, 81.43\% for for $2 \times 2$, 81.88\% for $4 \times 4$ and 80.83\% for $8 \times 8$. We thus observe that performance improves with spatial resolution up to $4 \times 4$, which the optimal, and then drops at $8 \times 8$. This drop may be due to assuming the same target at each spatial position. The resolution $8 \times 8$ is also more expensive computationally.


\end{document}